%% file: iclr2027_conference.tex
\documentclass{article} 
\usepackage{iclr2027_conference,times}
\input{math_commands.tex}

\usepackage{hyperref}
\usepackage{url}
\usepackage{booktabs}
\usepackage{colortbl}
\usepackage{xcolor}
\usepackage{graphicx}
\usepackage{tabularx}
\usepackage{rotating}
\usepackage{tikz}
\usepackage{geometry}
\usetikzlibrary{calc}
\usepackage{amssymb}
\usepackage{caption}
\usepackage{enumitem}
\usepackage{wrapfig}
\usepackage{array}
\usepackage{listings}
\usepackage{longtable}
\definecolor{tabhi}{HTML}{2F6F9F}
\definecolor{icdark}{HTML}{2F6F9F}
\definecolor{icmid}{HTML}{86AECB}
\definecolor{iclite}{HTML}{CFE0EC}
\definecolor{icgrey}{HTML}{8A99A5}

\definecolor{lvqaDown}{RGB}{202,74,56}
\definecolor{lvqaUp}{RGB}{47,124,140}
\definecolor{lvqaBase}{RGB}{120,120,120}

\newcommand{\rot}[1]{\rotatebox{90}{\scriptsize #1}}
\newcommand{\stk}[2]{%
    \begin{tabular}[c]{@{}c@{}}
        #1\\[-3.5pt]
        {\tiny\textcolor{black!55}{$\pm$#2}}
    \end{tabular}%
}
\newcolumntype{Z}{>{\centering\arraybackslash}m{7.2mm}}

\newcommand{\grp}[1]{%
    \makebox[0pt][c]{\textsc{#1}}%
}
\providecommand{\nb}[1]{\makebox[8.4mm][c]{#1}}
\definecolor{promptbg}{RGB}{248,248,246}
\lstdefinestyle{prompt}{
    basicstyle=\ttfamily\scriptsize, backgroundcolor=\color{promptbg},
    frame=leftline, framerule=1.2pt, rulecolor=\color{lvqaUp},
    framexleftmargin=6pt, xleftmargin=8pt, breaklines=true,
    breakindent=0pt, columns=fullflexible, keepspaces=true,
    showstringspaces=false, aboveskip=6pt, belowskip=6pt}

\newlength{\lvqatile}
\newcommand{\lvqachip}[1]{{\setlength{\fboxsep}{1.6pt}\colorbox{lvqaUp!14}{\scriptsize\sffamily\textcolor{lvqaUp!75!black}{#1}}}}

\title{Composition, Not Conversation: \\ VLMs Lose the Scene, Not the Thread}

\vspace{-15pt}

\author{
L D M S Sai Teja\textsuperscript{1,*},
Ufaq Khan\textsuperscript{2,*},
N Siva Gopala Krishna\textsuperscript{3,*}
Satyajit Tourani\textsuperscript{2},
\\
\textbf{Ashshak Sharifdeen\textsuperscript{2},}
\textbf{Fida Mohammad Thoker\textsuperscript{4},}
\textbf{Bernard Ghanem\textsuperscript{4},}
\textbf{Muhammad Haris Khan\textsuperscript{2}}
\\
\footnotesize
\textsuperscript{1}NIT Silchar
\quad
\textsuperscript{2}MBZUAI
\quad
\textsuperscript{3}BML Munjal University
\quad
\textsuperscript{4}KAUST
\quad\\
\textsuperscript{*}Equal contribution,
\quad
\href{https://github.com/lost-in-layers/composition-not-conversation}{
    \raisebox{-0.15\height}{\includegraphics[height=1.0em]{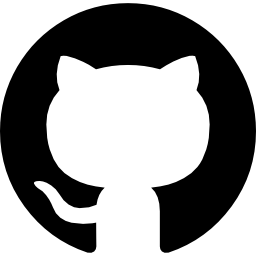}}
    \,\texttt{composition-not-conversation}
}
\quad
\href{https://huggingface.co/lost-in-layer/Layered-VQA}{
    \raisebox{-0.15\height}{\includegraphics[height=1.0em]{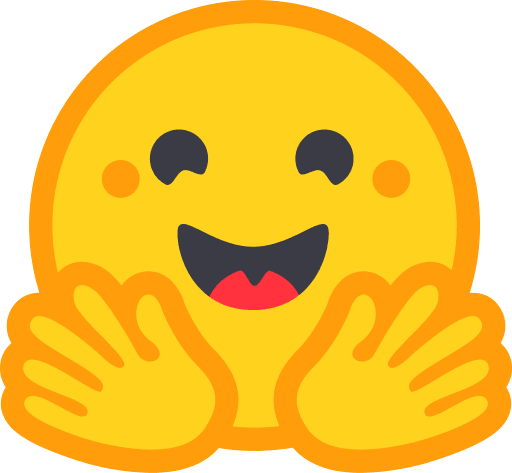}}
    \,\texttt{Layered-VQA}
}
}

\iclrfinalcopy 
\begin{document}

\maketitle

\vspace{-23pt}

\begin{figure}[h]
    \centering
    \includegraphics[width=1.0\linewidth]{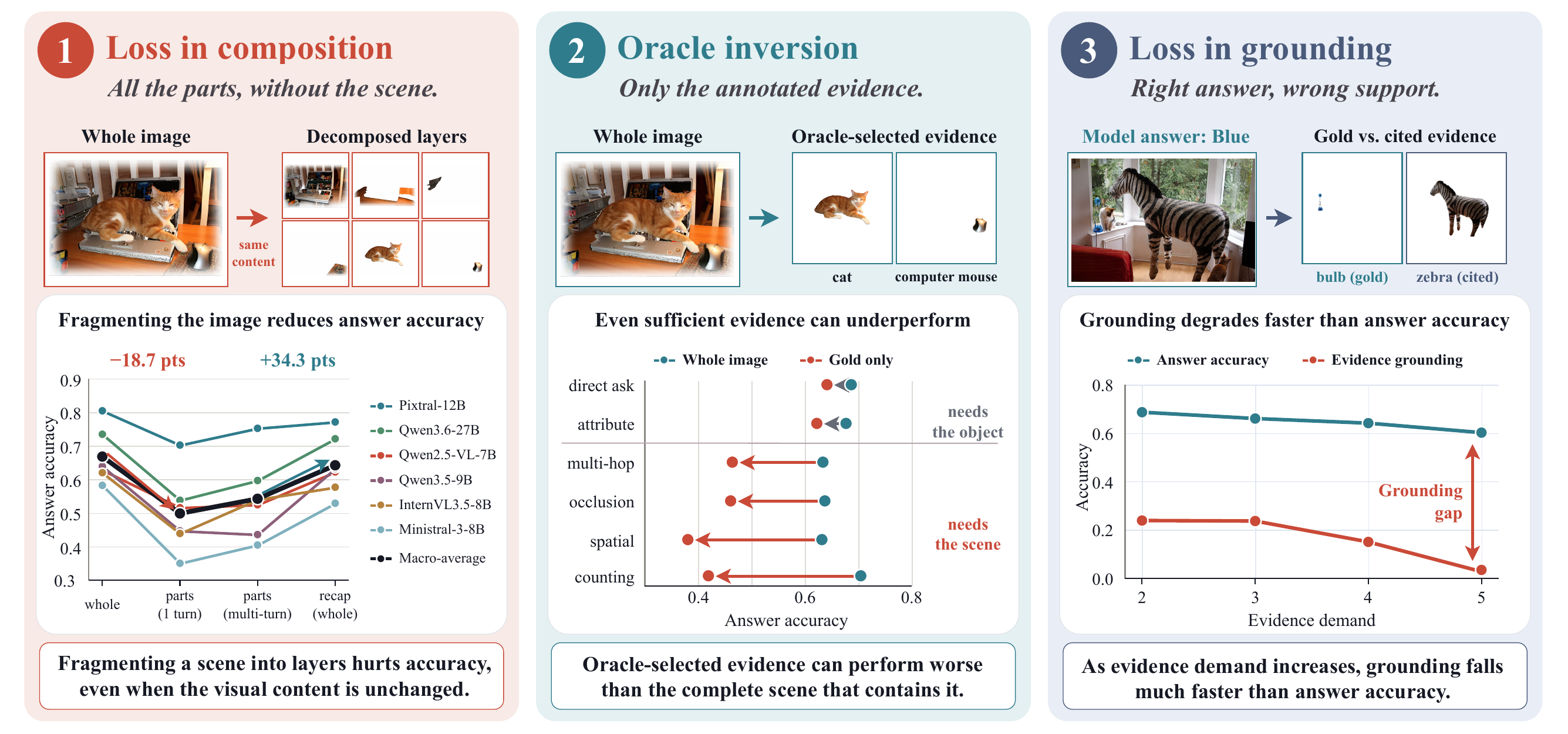}
\caption{
\textbf{Vision-language models lose the scene, not the thread.}
\textbf{\textit{Left:}} Scene fragmentation hurts accuracy, while recomposition recovers it.
\textbf{\textit{Middle:}} Oracle-selected minimal evidence can perform worse than the complete scene.
\textbf{\textit{Right:}} As evidence demand grows, grounding falls much faster than answer accuracy.
}
    \label{fig:main-teaser}
    \vspace{-7pt}
\end{figure}

\begin{abstract}
Vision-language models (VLMs) increasingly reason over visual evidence that is cropped, segmented, retrieved, or revealed over time.
Yet most VQA benchmarks present the complete image and question at once.
We ask what models lose when the same information is fragmented.
We introduce \textbf{Layered-VQA}, with 93 scenes and 300 questions.
% We introduce \textbf{Layered-VQA}\footnote{Code on \href{https://github.com/lost-in-layers/composition-not-conversation}{GitHub}, Data on \href{https://huggingface.co/lost-in-layer/Layered-VQA}{Hugging Face}.}, with 93 scenes and 300 questions.
Each image is decomposed into ordered RGBA layers that exactly recompose the original scene, and each question is annotated with supporting, minimal-sufficient, and distractor layers.
We evaluate eleven open-weight VLMs from $3$B to $32$B parameters and two proprietary models with a scale of $187{,}200$ conversations, graded by $1.74M$ open-model cross-judgments.
We find three consistent failures.
\textbf{\textit{Loss in Composition:}} fragmenting the question has a small effect, but fragmenting the scene substantially reduces accuracy; recomposing the same layers largely restores performance.
\textbf{\textit{Oracle Inversion:}} even oracle-selected sufficient evidence can perform worse than the complete scene.
\textbf{\textit{Loss in Grounding:}} as more evidence is required, grounding degrades much faster than answer accuracy.
Together, these results show that having the right visual evidence is not enough.
How that evidence is composed and presented determines whether models can use and ground it.
\textbf{\emph{The right evidence is not enough: VLMs need the scene it came from.}}
\end{abstract}

\input{sections/1-introduction-fida}
\input{sections/2-related-work}
\input{sections/3-layered-vqa}
\input{sections/4-conv-smi}
\input{sections/5-comp-not-conv}
\input{sections/6-conclusion}
\input{sections/ethics-reproduce-ai}
\clearpage

\bibliography{iclr2027_conference}
\bibliographystyle{iclr2027_conference}

\clearpage
\appendix
\input{sections/appendix-additional-0}
\input{sections/appendix-additional-1}
\clearpage
\input{sections/appendix-example}
\clearpage
\input{tables/appendix-data-preview}
\clearpage
\input{sections/appendix-prompts}

\end{document}

%% file: math_commands.tex
\usepackage{amsmath,amsfonts,bm}

\def\eqref#1{equation~\ref{#1}}
\def\1{\bm{1}}

\DeclareMathAlphabet{\mathsfit}{\encodingdefault}{\sfdefault}{m}{sl}
\SetMathAlphabet{\mathsfit}{bold}{\encodingdefault}{\sfdefault}{bx}{n}

%% file: sections/1-introduction-fida.tex
\section{Introduction}
% \vspace{-15pt}
The performance of a VLM may depend not only on what information is available but also on \textit{how} that information is presented.
This question is becoming increasingly relevant with the rise of multimodal agents, which reason and act over multiple steps while interacting with tools and evolving visual observations~\citep{koh-etal-2024-visualwebarena}.
In practical settings, visual evidence may be cropped, segmented, retrieved, or revealed incrementally~\citep{gupta2023visual,suris2023vipergpt,hu2023avis,liu2024llava}, while user queries may also unfold or be revised across turns~\citep{das2017visual,liu2024mmdu}.
This creates a growing mismatch between how VLMs are evaluated and how they increasingly operate in practice.
Existing visual question answering (VQA) and general-purpose VLM benchmarks typically evaluate models in a standard complete-input setting: one complete image, one complete question, and one turn~\citep{goyal2017making,hudson2019gqa,liu2024mmbench,yu2023mm}.
We therefore ask a simple question: \textit{How does the delivery of visual information affect VLM performance when the underlying information remains unchanged?}
In particular, we seek to understand whether the same visual information leads to different model behavior when its composition, delivery across turns, or alignment with the evolving query changes.

Recent work on language models provides an explanation: when a complete instruction is fragmented across turns, performance degrades as models commit prematurely and fail to revise their answers as the interaction unfolds~\citep{laban2026llms}. 
A similar effect may be expected for VLMs, particularly as visual evidence is increasingly distributed across long, multi-turn, or retrieval-based contexts~\citep{song2024milebench,wang2025multimodal,wang2024needle,wu2025visual}. 
Yet visual information introduces an additional source of complexity: distributing an image does not only change when information arrives, but can also break the composition of the scene itself. 
Consequently, a drop in VLM performance can have several explanations: the model may struggle with information arriving across turns, with integrating fragmented visual content into a coherent scene, or with information that is lost during evidence selection. 
Existing evaluations do not disentangle these effects, making it difficult to determine whether the primary challenge lies in the conversation, the visual composition, or the evidence available to the model. 
Isolating these factors therefore requires a setting in which the underlying visual information is controlled while only its form and delivery are varied.

To isolate these factors, we introduce \textbf{\textit{Layered-VQA}}, a controlled benchmark in which the same underlying visual information can be presented in different forms.
Each scene consists of a composite image and an ordered set of RGBA layers that exactly recompose the original scene~\citep{tudosiu2024mulan}.
We manually author every question and annotate the layers that carry the answer, the smallest set of layers sufficient to answer it, and plausible but irrelevant layers.
This structure allows us to systematically vary how the same visual information is delivered: as a complete scene or fragmented into layers, within a single turn or across multiple turns, and aligned or misaligned with the question components it supports.
We additionally recompose fragmented evidence and construct oracle conditions containing only the minimally sufficient layers, allowing us to separate the effects of visual composition, conversational delivery, and evidence selection.
Across these controlled conditions, the underlying scene and question remain unchanged; what changes is how the information is presented to the model.

Our experiments reveal three main findings, Figure~\ref{fig:main-teaser}.
First, we find a clear \textbf{\textit{Loss in Composition}}: fragmenting the question has relatively little and inconsistent effect, whereas fragmenting the visual scene into layers consistently degrades performance.
This loss already emerges when all layers are presented together in a single turn, becomes larger when the same layers are distributed across turns, and is largely recovered when the layers are recomposed into the original image.
These results separate the cost of visual fragmentation from the additional cost of multi-turn interaction, pointing to scene composition as a central bottleneck.
Second, we uncover an \textbf{\textit{Oracle Inversion}}: even when models are given only the oracle-selected minimal set of layers sufficient to answer the question, they perform worse than when given the complete scene.
Thus, removing seemingly unnecessary visual content can make reasoning harder, indicating that surrounding scene structure contributes to how VLMs interpret otherwise sufficient evidence.
Finally, we observe a \textbf{\textit{Loss in Grounding}}: models lose the ability to identify the visual evidence supporting their predictions much faster than they lose answer accuracy.
Correct answers can therefore mask substantial failures in evidence attribution, revealing a gap between producing the right answer and producing it for the right visual evidence.

%% file: sections/2-related-work.tex
\section{Related Work}

\paragraph{\textsc{Multi-turn Interaction and Visual Composition:}}
Multi-turn benchmarks ask whether models can carry information across an interaction.
In text, \citet{laban2026llms} show that splitting one instruction across turns can sharply reduce performance.
\textit{Visual Dialog} keeps one image while questions unfold through dialogue history \citep{das2017visual}, \textit{ConvBench} chains perception, reasoning, and creativity \citep{liu2024convbench}, and \textit{MMDU} extends this setting to long multi-image conversations \citep{liu2024mmdu}.
At longer contexts, \textit{MileBench} measures reasoning over many images \citep{song2024milebench}, while \textit{MM-NIAH}, \textit{MMNeedle}, and \textit{Visual Haystacks} place relevant visual evidence among increasingly large or distracting contexts \citep{wang2024needle,wang2025multimodal,wu2025visual}.
These works vary dialogue history, context length, image count, or distractors, where as Layered-VQA preserves the source scene content and object coordinates, and varies only composition and delivery across turns.

Compositional benchmarks ask whether models bind objects, attributes, and relations correctly, Layered-VQA instead asks whether these relations survive scene fragmentation.
\textit{CLEVR} and \textit{GQA} use structured scenes and questions to diagnose compositional reasoning \citep{johnson2017clevr,hudson2019gqa}.
\textit{Winoground} tests relations through matched image-caption pairs \citep{thrush2022winoground}, \textit{ARO} tests attribution, relations, and word order \citep{yuksekgonul2023when}, and \textit{SugarCrepe} builds harder contrastive negatives while reducing linguistic shortcuts \citep{hsieh2023sugarcrepe}.
\citet{campbell2024understanding} further connect multi-object failures to feature-binding interference.
Layered-VQA removes only the composed view, asking whether models can recover the scene when the same visual content is supplied as separate layers.

\paragraph{\textsc{Decomposed Evidence and Grounding:}}
Image decomposition and visual tools change what evidence a model receives.
\textit{SAM} produces object masks \citep{kirillov2023segment}, \textit{pix2gestalt} completes partially hidden objects \citep{ozguroglu2024pix2gestalt}, \textit{MuLAn} provides ordered RGBA decompositions for controllable image generation \citep{tudosiu2024mulan}, and Transparent Image Layer Diffusion generates images as consistent transparent layers \citep{zhang2024transparent}.
Layered-VQA uses the layered scenes from \textit{MuLAn}, manually authors questions over them, and instead asks what visual reasoning is lost when these components replace the composed scene.  
Other work selects evidence during reasoning.
\textit{VisProg} and \textit{ViperGPT} execute modular visual programs \citep{gupta2023visual,suris2023vipergpt}, \textit{AVIS} and V$^*$ search for useful visual information \citep{hu2023avis,wu2024vstar}, and \textit{MuRAG} retrieves relevant multimodal evidence from external memory \citep{chen2022murag}.
Layered-VQA does not test whether a model can discover evidence.
\textit{The evidence is fixed and annotated, allowing us to control for retrieval as a potential confound and ask a different question: once the relevant visual evidence is available, does selecting or separating it preserve the reasoning supported by the complete scene?}

This distinction also matters for grounding.
\citet{jacovi2020towards} separate plausible explanations from faithful ones, \textit{ERASER} evaluates whether annotated rationales are sufficient and necessary \citep{deyoung2020eraser}, and \citet{turpin2023language} show that model explanations can omit factors that changed the answer.
In VQA, \citet{reich2024uncovering} show that the benefits of visual grounding methods are underestimated because visual features frequently omit the question-relevant information such methods rely on. \textit{HallusionBench} diagnoses visual illusion and language hallucination through controlled question pairs \citep{guan2024hallusionbench}.
Layered-VQA makes the supporting visual evidence explicit.
Each question has gold, minimal-sufficient, and distractor layers, and models must cite the layers supporting their answer.
This exposes a failure that answer accuracy alone cannot show: a model can give the right answer while citing the wrong visual evidence, without implying that the cited layers faithfully reflect its internal reasoning.

%% file: sections/3-layered-vqa.tex
\section{Layered-VQA}
\label{sec:layered_vqa}

\begin{figure}[h]
    \centering
    \includegraphics[width=1.0\linewidth]{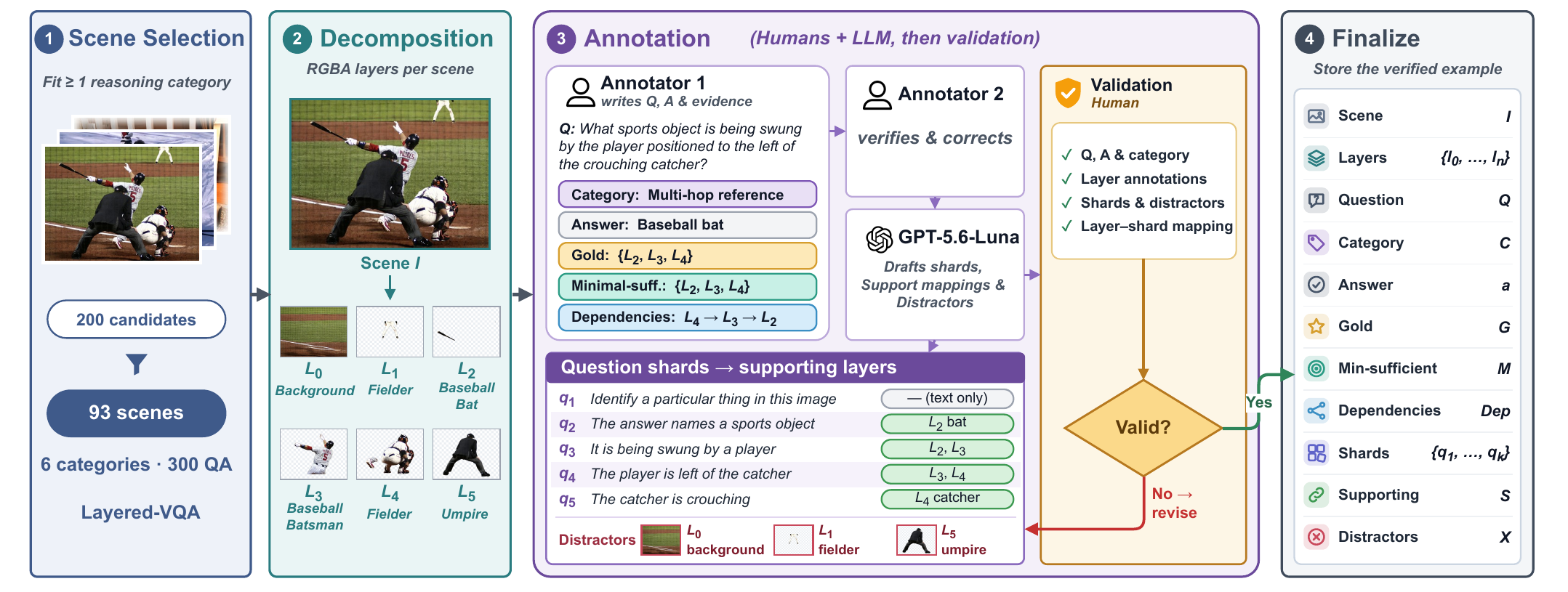}
    \caption{\textbf{Layered-VQA construction.}
    Scenes are decomposed into RGBA layers, manually annotated, expanded into question shards and evidence mappings with GPT-5.6-Luna, and verified before finalization.}
    \label{fig:benchmark}
\end{figure}

Layered-VQA turns each VQA example into a fixed set of visual and linguistic components that can be presented in different forms.
Each example contains a composite scene $I$, its RGBA layers $\mathcal{L}$, a question $Q$, an answer $a$, question shards $\mathcal{Q}$, and explicit evidence annotations.
The scene, question, and answer stay fixed across conditions.
Only how they are composed and delivered changes.
We begin with 200 candidate scenes, retain 93 that support at least one reasoning category, and construct 300 questions across six categories. The entire benchmark preparation is given in Figure~\ref{fig:benchmark}.

\paragraph{Layered Scenes:}
Each scene comes from MuLAn~\citep{tudosiu2024mulan} and contains a composite image $I$ with an ordered stack of RGBA layers $\mathcal{L}=\{L_0,\ldots,L_n\}$.
MuLAn layers are object instances on the full canvas.
The layers exactly recompose the original scene.
$L_0$ contains the background, while the remaining layers contain individual objects on the original full-size canvas.
The source scene and canvas coordinates therefore remain fixed when moving between the composite and its layers.
This gives two views of the same scene: the composed image $I$ and its decomposed representation $\mathcal{L}$.

\paragraph{Questions, Evidence, and Verification:}
We write questions in six categories:
\textit{Direct ask}, \textit{Attribute binding}, \textit{Counting}, \textit{Spatial}, \textit{Occlusion}, and \textit{Multi-hop reference}.
For each scene, Annotator 1 writes the category $C$, complete question $Q$, answer $a$, gold evidence $G$, minimal-sufficient evidence $M$, and evidence dependencies $\mathrm{Dep}$.
$G$ contains the layers supporting the complete question.
$M$ is the smallest layer set from which a person can still determine the answer.
$\mathrm{Dep}$ records dependencies between evidence layers.
Annotator 2 checks all of these annotations and corrects disagreements after discussion.
GPT-5.6-Luna then decomposes the verified question into ordered shards
$\mathcal{Q}=\{q_1,\ldots,q_m\}$.
For each shard $q_i$, it also assigns the supporting layer set $S_i$ and proposes distractor layers $X$.
A shard may be text-only, in which case $S_i=\varnothing$.
Together, the shards preserve the information in the complete question, while the support mapping makes the correspondence between each question part and its visual evidence explicit.
Distractors belong to the same scene but are not required to determine the answer.
The complete annotation is then checked by human.
Validation covers the question, answer, and category; the gold, minimal-sufficient, and dependency annotations; the shards and distractors; and every shard-to-layer mapping.
Incorrect annotations are revised and checked again before the sample is finalized.
Each verified example stores $(I,\mathcal{L},Q,C,a,G,M,\mathrm{Dep},\mathcal{Q},S,X)$.
When layers are shown to a model, they are referred to only by numerical indices, so semantic layer names cannot reveal the answer.
Layered-VQA contains 83 \textit{Direct ask}, 65 \textit{Attribute binding}, 54 \textit{Spatial}, 44 \textit{Counting}, 37 \textit{Multi-hop reference}, and 17 \textit{Occlusion} questions.
Questions contain 3 to 6 shards, most commonly 5, and require 1 to 6 gold layers.
Every question contains at least one distractor, and 197 questions contain an evidence dependency.
These annotations let the same VQA problem be evaluated under different forms of composition, timing, evidence order, and question-to-evidence alignment.

%% file: sections/4-conv-smi.tex
\section{Experimental Setup}
\label{sec:experimental_setup}

Starting from the Layered-VQA representation introduced in Section~\ref{sec:layered_vqa}, we construct controlled conditions that vary how the question shards and visual layers are presented to a VLM.
Our experiments are designed to separate three factors: whether the visual scene is composed or fragmented, whether information is delivered in one turn or across multiple turns, and which visual evidence is available to the model.

\paragraph{\textsc{Controlled Delivery Conditions:}}
% \subsection{Controlled Delivery Conditions}
\label{sec:delivery_conditions}

\begin{wrapfigure}{r}{0.65\textwidth}
    \centering
    \vspace{-10pt}
    \includegraphics[width=\linewidth]{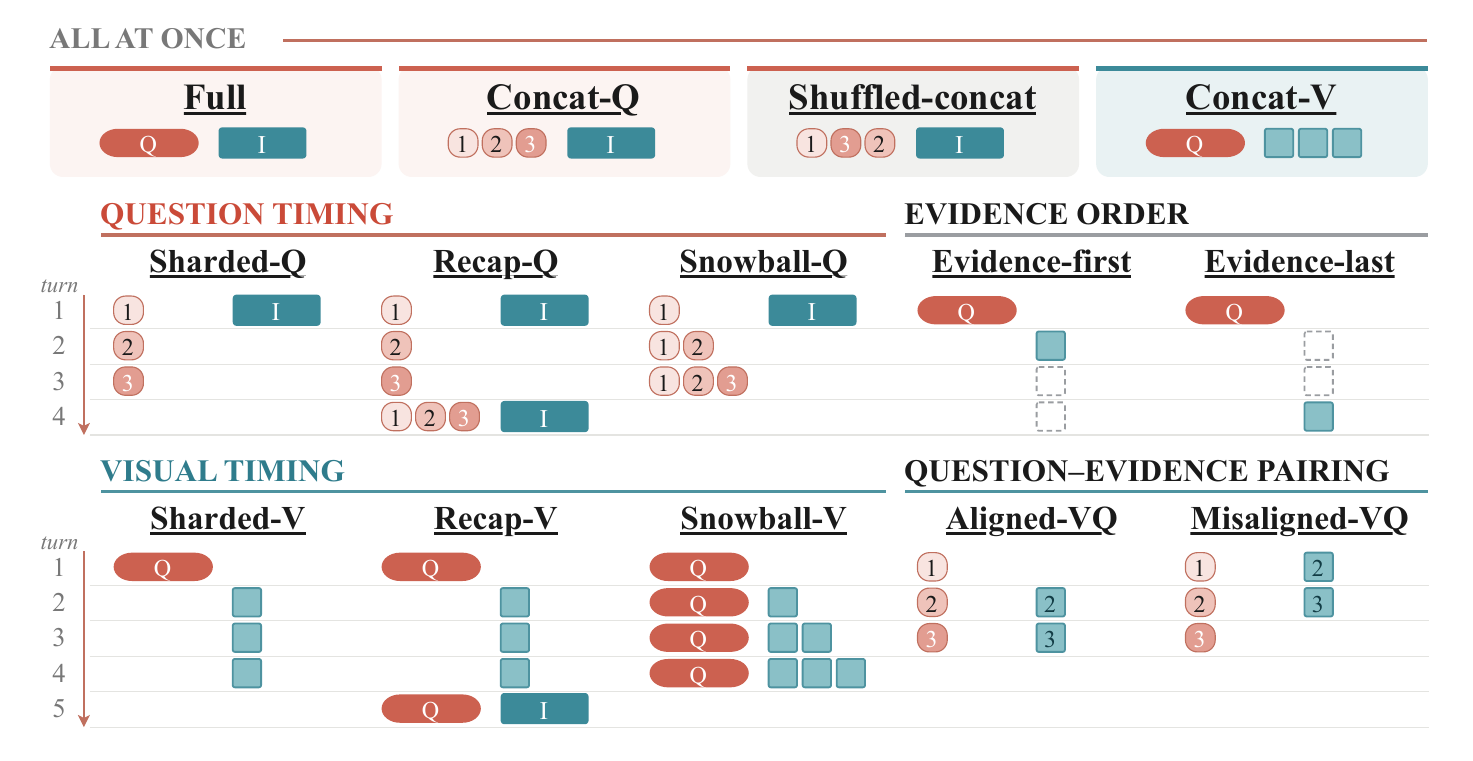}
    \caption{\textbf{Controlled delivery conditions.}
    The sample is presented under different question, visual, alignment, and evidence schedules.}
    \label{fig:sim-conditions}
    \vspace{-10pt}
\end{wrapfigure}

We evaluate sixteen conditions, summarized in Figure~\ref{fig:sim-conditions}.
Fourteen preserve the complete question and visual content and change only  how they are composed or delivered.
The remaining two, \textsc{all-layers} and \textsc{oracle}, modify the available evidence for grounding and oracle-evidence evaluation.
We organize the conditions according to which component of the Layered-VQA representation is manipulated.

\textbf{\textsc{Question delivery:}}
The 1\textsuperscript{st} family varies how the question is presented while keeping the
composite scene available.
\textsc{\textbf{full}} presents the complete question $Q$ and composite image $I$ in one turn.
\textsc{\textbf{concat-q}} presents all question shards together, while \textsc{\textbf{shuffled-concat}} presents the same shards in random order.
\textsc{\textbf{sharded-q}} reveals one shard per turn.
\textsc{\textbf{recap-q}} ends the sharded conversation by restating the complete question.
\textsc{\textbf{snowball-q}} repeats all previous shards whenever a new shard arrives.
\textbf{\textsc{Visual delivery:}}
The 2\textsuperscript{nd} family keeps the complete question fixed and varies the
presentation of the visual scene.
\textsc{\textbf{concat-v}} presents every layer together without the composite image.
\textsc{\textbf{sharded-v}} reveals one layer per turn.
\textsc{\textbf{recap-v}} ends the sharded sequence by showing the recomposed scene.
\textsc{\textbf{snowball-v}} repeats all previously revealed layers at every turn.
\textsc{\textbf{evidence-first}} and \textsc{\textbf{evidence-last}} keep the same layers but place the annotated evidence early or late.

\textbf{\textsc{Joint Question-visual alignment:}}
The 3\textsuperscript{rd} family fragments both streams.
\textsc{\textbf{aligned-vq}} pairs each question shard with its annotated supporting layers introduced in Section~\ref{sec:layered_vqa}.
\textsc{\textbf{misaligned-vq}} preserves the same layers and per-turn layer counts but pairs each shard with layers that do not support it.
This isolates whether local question-evidence alignment can compensate for a fragmented scene.

\textbf{\textsc{Evidence intervention:}}
The final family changes which representation of the visual evidence is available while keeping the complete question in a single turn.
\textsc{\textbf{all-layers}} presents the composite image together with every layer and asks the model to identify the layers supporting its answer.
\textsc{\textbf{oracle}} removes the composite and distractors and presents only the annotated minimal-sufficient layers.
These two measures: grounding when all evidence is available and can the annotated evidence replace the complete scene.
% The first measures grounding when all evidence is available.
% The second asks whether exactly the annotated evidence can replace the complete scene.

\paragraph{\textsc{Multi-turn Simulation:}}
% \subsection{Multi-turn Simulation}
\label{sec:simulation}

For multi-turn conditions, we simulate an interaction between a user agent and the VLM being evaluated.
The user agent holds the unrevealed question shards and the condition-specific delivery schedule, while the VLM has access only to the information revealed up to the current turn.
The VLM sees only the information revealed so far, and each response is appended to the conversation history.
The interaction continues until all scheduled information has been delivered, even if the model answers correctly earlier.
The assistant is told that the question and visual evidence may arrive across turns and that each layer is part of a decomposed scene.
We use two user agents.
The \textit{scripted user} follows the fixed benchmark schedule.
The \textit{LLM user} selects a remaining question shard and rephrases it conversationally while preserving its information.
It could reorder the layers, which would change the schedule under test.
It uses the same model as the assistant at temperature $1.0$ and forbids adding new facts, describing the image, or answering the question.
The LLM user is used only for \textsc{sharded-q}, \textsc{recap-q}, and \textsc{snowball-q}.
Conditions with fragmented visual input use the scripted user to preserve the layer order and evidence schedule.
Single-turn conditions use no user agent.
% We do not apply a separate semantic-equivalence check after rephrasing.

\paragraph{\textsc{Evaluation Protocol:}}
\label{sec:evaluation_protocol}

\begin{wrapfigure}{r}{0.62\textwidth}
    \centering
    \vspace{-6pt}
    \includegraphics[width=\linewidth]{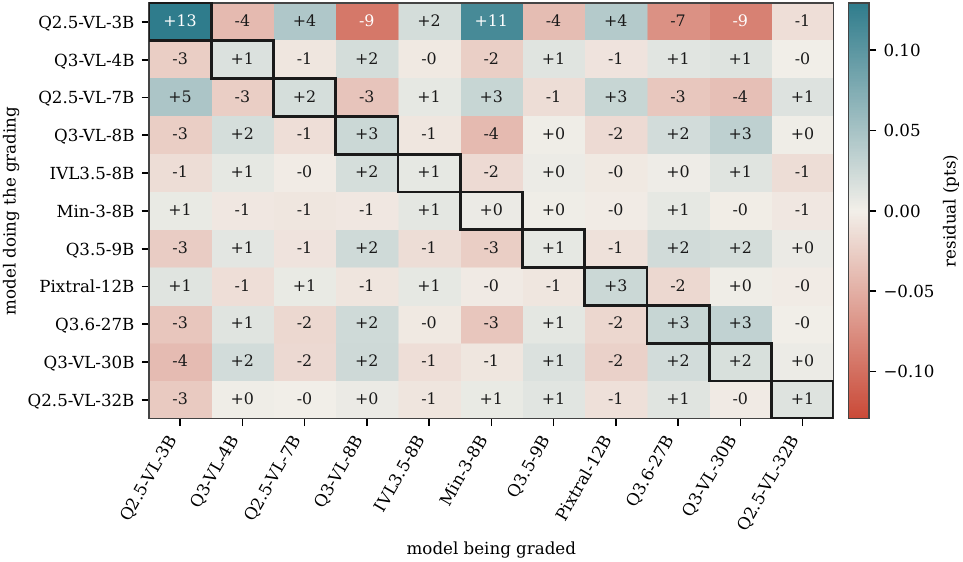}
    \caption{\textbf{Self-preference after cross-judging.}
    Boxed cells correspond to models judging their own outputs.}
    \label{fig:selfpref}
    \vspace{-10pt}
\end{wrapfigure}

\textbf{\textsc{Answer accuracy:}}
Answers are free-form, so completed conversations are evaluated by an LLM judge against the reference answer. 
We score the assistant's final response against the reference answer.
Earlier correct answers do not count if the model later changes or withdraws them.
This gives every condition one final prediction regardless of conversation length.
Every completed conversation is independently graded by all eleven open models.
Our primary metric is \textbf{Consensus Final-turn Accuracy}, obtained by majority vote across the eleven judges.
\textsc{retained} is a model's mean accuracy across a family's conditions, as a percentage of its own \textsc{full} score.
On $400$ conversations, the consensus matches humans 96.0\% of the time ($\kappa = 0.92$), against 92.5\% for the best single judge, more in Appendix~\ref{app:human}.

\textbf{\textsc{Grounding:}}
In \textsc{all-layers}, the model also reports the layers supporting its answer.
We compare this set with the annotated evidence using precision, recall, F1, exact match, completeness, excess rate, distractor rate, and invalid rate.
\textbf{Joint accuracy} requires both the answer and the predicted layer set to be correct.
The \textbf{grounding gap} is the difference between answer accuracy and joint accuracy.

\textbf{\textsc{Cross-Judging and self-preference:}}
In the primary evaluation, each open model judges its own conversations.
To quantify potential self-preference, we additionally re-grade every
conversation with every open-model judge.
The \textit{self-preference residual} measures the remaining self-grading effect after accounting for assistant ability and judge leniency.
This analysis is diagnostic, shown in Figure~\ref{fig:selfpref}.
Every model favours itself, by 2.6 points on average and by 12.9 for the smallest, so a single self-judge lifts exactly the models a benchmark needs to separate.

\noindent \textbf{\textsc{Uncertainty:}}
Every condition is run three times.
Reported values are the mean $\pm$ standard deviation across these three runs.
For the main paired comparisons, we additionally report $95\%$ confidence intervals from a scene-level bootstrap, resampling scenes rather than individual questions.

\subsection{Models and Implementation}
\label{sec:models_implementation}
\vspace{-2mm}
We evaluate eleven open-weight VLMs from $3$B to $32$B parameters: Qwen2.5-VL-\{3B, 7B, 32B\}~\cite{bai2025qwen25vltechnicalreport}, Qwen3-VL-\{4B, 8B, 30B-A3B\}~\cite{bai2025qwen3vltechnicalreport}, InternVL3.5-8B~\cite{wang2025internvl35advancingopensourcemultimodal}, Ministral-3-8B~\cite{liu2026ministral3}, Qwen3.5-9B~\cite{qwen3.5}, Pixtral-12B~\cite{agrawal2024pixtral12b}, and Qwen3.6-27B~\cite{qwen3.6-27b}.
We additionally evaluate GPT-5.6-Luna and Gemini-3.7-Flash as proprietary assistants.
Every model is evaluated on all sixteen conditions with three independent runs.

Open models are served with vLLM~\cite{kwon2023efficient} on two AMD MI210 $64$GB GPUs.
Models up to $27$B use one GPU, while the 30B and 32B models use tensor parallelism across both.
The proprietary models are accessed through OpenRouter.
All generations use temperature $1.0$, $\text{top-}p=1.0$, no top-$k$ truncation, and at most $1024$ output tokens.
The context window is $65{,}536$ tokens where supported and otherwise the largest available value, always at least $32{,}768$.
Across all thirteen assistants, the evaluation contains $187{,}200$ conversations: $13$ models $\times$ $16$ conditions $\times$ $300$ questions $\times$ $3$ runs.
The 158,400 conversations generated by the 11 open models are each graded by all 11 open-model judges, giving $1.74M$ cross-judgments.

%% file: sections/5-comp-not-conv.tex
\vspace{-3mm}
\section{Results: Composition, not Conversation}
\vspace{-3mm}
We separate \emph{composition} from \emph{conversation}: whether the scene is whole or decomposed, and whether information arrives in one turn or across several.
The question and source scene remain fixed; only presentation and delivery change.
\vspace{-3mm}

\input{tables/Exp-B}

\paragraph{\textsc{Loss in composition:}}
\label{sec:composition}

\textbf{\textit{A scene is more than the layers it contains.}}
Replacing the whole image with all of its layers in the same turn drops accuracy by $-.187$ across all models.
No visual evidence is removed, only the composition is lost.
Doing the same to the question costs just $-.041$.
Text survives fragmentation much better than vision.
\textbf{\textit{Memory alone does not explain this gap.}}
Spreading either question fragments or image layers across turns adds a similar penalty, $-.103$ and $-.101$.
In \textsc{snowball}, every turn repeats all previously seen layers, so the model no longer needs to remember them.
Performance returns almost exactly to the all-layers-in-one-turn condition, but not to the whole-image condition.
The model can hold all the parts without recovering the scene.
\textbf{\textit{Recomposition does.}}
In \textsc{recap}, the same sharded conversation ends by showing the whole image once.
Yet accuracy rises by $+.343$ ($95\%$ CI $[+.315, +.373]$), with every model improving.
The recovered score even exceeds showing the whole image only at the start.
What matters is not simply whether the model has seen the evidence, but whether that evidence is available as a composed scene when it answers.
\textbf{\textit{Visual evidence also has to arrive when it is useful.}}
The same layers score $.410$ when useful evidence arrives late, but only $.265$ when it arrives before the question.
Visual evidence is therefore substantially less useful when it arrives before the question.
The model does not reliably hold it in reserve until the question gives it meaning.
Even where alignment should help, it does not: relational questions score $.313$ in both conditions, and aligned pairing underperforms misaligned for 9/11 models.
As Table~\ref{tab:exp-b} shows, the pattern is consistent: local alignment does not recover the scene, question fragmentation is relatively cheap, visual fragmentation is costly, and combining both is worse.
\textbf{\textit{Visual evidence is not additive.}}
Having all the parts is not equivalent to seeing the scene.
Models reason best when visual evidence is composed and available at the moment it is needed.

\paragraph{\textsc{Oracle inversion:}}
\label{sec:oracle-inversion}

\input{tables/Exp-A}
\textbf{\textit{Perfect retrieval makes the model worse.}}
Table~\ref{tab:exp-a} presents the same visual evidence in three ways.
\textsc{rgb} shows the whole image.
\textsc{all-layers} shows the image together with every layer.
\textsc{oracle} keeps only the annotated layers needed to answer the question.
Oracle removes irrelevant content while preserving everything required for the answer.
Instead, accuracy falls by $-.148$ relative to the whole image, and this happens for all models.
Giving the model exactly the annotated evidence is worse than giving it the scene.
\textbf{\textit{The layer format does not explain the loss.}}
\textsc{all-layers} contains the same layer renders used by \textsc{oracle}, but keeps the original image beside them.
Its change from \textsc{rgb} is only $-.017$.
The important difference is what \textsc{oracle} removes.
It keeps the evidence directly tied to the answer, but removes the rest of the scene.
\textbf{\textit{Relevant evidence is not sufficient evidence.}}
The model does not reason from isolated answer-bearing regions in the same way that it reasons from a complete scene.
Background, surrounding objects, and global composition can help interpret evidence that looks sufficient on its own.
A retrieval step can therefore become destructive even when it retrieves the correct regions perfectly.
Better visual retrieval is not only about finding the right evidence.
It is also about preserving the context that makes that evidence meaningful.

\paragraph{\textsc{Loss in grounding:}}
\label{sec:grounding}

\textbf{\textit{A correct answer does not mean the model used the right evidence.}}
In \textsc{all-layers}, models answer with $.587$ accuracy, but answer and grounding are jointly correct only $.090$ of the time.
Their cited evidence is often not merely incomplete.
Models over-cite layers, select distractors, and sometimes cite layers that do not exist.
A model can therefore reach the right answer while giving a poor account of where that answer came from.
\textbf{\textit{Answering and grounding are different abilities.}}
The model ranking changes sharply when evidence is scored.
Qwen3-VL-4B performs well, but poor grounding, answers .693 of questions with an evidence F1 of .346, while Qwen3.6-27B is strong in both, answers .736 with an F1 of .654.
Answer accuracy alone hides this difference.
Two models can look equally capable while reasoning from the image very differently.
\textbf{\textit{The conversation reveals an even stronger failure.}}
In the sharded setting, we know exactly when each supporting layer becomes available.
If the model answers before its gold evidence arrives, that answer cannot be grounded in that evidence.
We call such cases \emph{ungrounded correct}.
\begin{wrapfigure}{r}{0.50\linewidth}
    \centering
    \vspace{-5mm}
    \includegraphics[width=\linewidth]{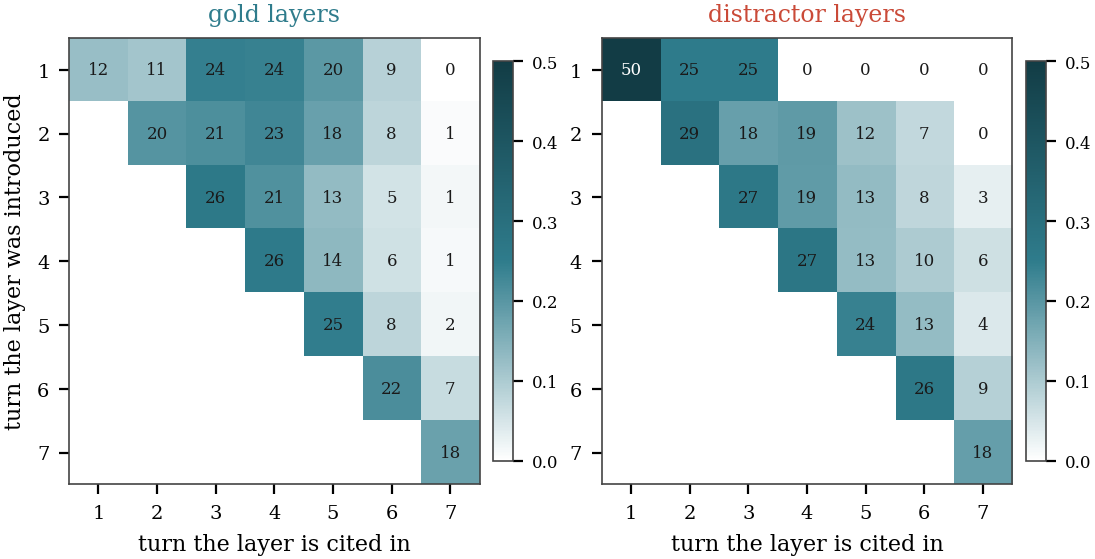}
    \caption{\textbf{Layer citation over time.}}
    \label{fig:citations}
    \vspace{-5mm}
\end{wrapfigure}
Across the visual conditions, $.393$ of correct answers arrive before the required evidence.
Requiring the evidence to be present before the answer reduces accuracy from $.420$ to $.270$.
Much of what standard accuracy rewards is therefore not supported by the visual evidence available at the time.
\textbf{\textit{This changes how visual robustness should be read.}}
A model that can answer without seeing the required evidence may appear robust when the image is fragmented, delayed, or degraded.
Its score survives partly because its answer does not depend on the manipulated evidence.
\textbf{\textit{Answer accuracy measures whether the model is right.
Grounding measures whether it is right for the image.}}

\paragraph{\textsc{How the answers are produced:}}

\textbf{\textit{Answering early is worse than answering late.}}
Accuracy by the point at which a model first commits runs $.388$, $.458$, $.460$, $.473$, $.403$ across fifths of the conversation, Table~\ref{tab:first-answer-attempt}.
Committing in the first fifth is the worst option, and all the models commit there most often.
\textbf{\textit{Longer replies are worse replies.}}
Within each model's own length distribution, the longest fifth of responses is the least accurate, in both single-turn and multi-turn conditions, Table~\ref{tab:verbosity}.
The drop from shortest to longest is negative for all models and reaches $-.193$.
\textbf{\textit{Being right is not the same as being grounded.}}
Scoring only the answers that arrived after their supporting layer takes the field from $.420$ to $.270$, Table~\ref{tab:ungrounded-grounded}.
More than a third of correct answers were produced before the evidence for them existed.
\textbf{\textit{A layer is cited near the turn it arrives, and less often as the conversation moves on.}}
Figure~\ref{fig:citations} plots where each layer is cited against the turn it arrived in.
Two fifths of all citations of a layer fall in the turn that delivered it, and two turns later the share is under a fifth.
A layer delivered in the sixth turn is cited in that turn $76\%$ of the time, against $12\%$ for one delivered first.
There is simply no conversation left to carry it, which is the mechanism \textsc{recap-v} exploits.
The scene is never new information; it is information the model was never given room to use.
Appendices~\ref{app:bloat}, \ref{app:reliability}, \ref{app:premature}, \ref{app:verbosity}, \ref{app:grounded}, \ref{app:judges} cover answer length, consistency, early commitment, verbosity, ungrounded answers, and judge self-preference.
% Appendix~\ref{app:bloat} shows that answer attempts get shorter rather than longer as a conversation runs, the opposite of the text-only result.
% Appendix~\ref{app:reliability} shows that sharding costs accuracy and buys inconsistency at the same time.

%% file: tables/Exp-B.tex
\begin{table*}[t]
\centering
\small
\vspace{-2mm}
\caption{\textbf{What it costs to deliver the evidence across turns.}
Consensus final-turn accuracy.
}
\label{tab:exp-b}
\renewcommand{\arraystretch}{1.05}

\resizebox{\textwidth}{!}{%
\begin{tabular}{l c cccc cccc cc ccc ccc}
\toprule
& \multicolumn{1}{c}{\textsc{\textbf{base}}}
& \multicolumn{4}{c}{\textsc{\textbf{question in text}}}
& \multicolumn{4}{c}{\textsc{\textbf{question in layers}}}
& \multicolumn{2}{c}{\textsc{\textbf{alignment}}}
& \multicolumn{3}{c}{\textsc{\textbf{order}}}
& \multicolumn{3}{c}{\textsc{\textbf{retained \%}}} \\

\cmidrule(lr){2-2}
\cmidrule(lr){3-6}
\cmidrule(lr){7-10}
\cmidrule(lr){11-12}
\cmidrule(lr){13-15}
\cmidrule(lr){16-18}

Model
& \rot{\textsc{\textbf{full}}}
& \rot{\textsc{\textbf{concat}}}
& \rot{\textsc{\textbf{sharded}}}
& \rot{\textsc{\textbf{recap}}}
& \rot{\textsc{\textbf{snowball}}}
& \rot{\textsc{\textbf{concat}}}
& \rot{\textsc{\textbf{sharded}}}
& \rot{\textsc{\textbf{recap}}}
& \rot{\textsc{\textbf{snowball}}}
& \rot{\textsc{\textbf{aligned}}}
& \rot{\textsc{\textbf{misaligned}}}
& \rot{\textsc{\textbf{ev-first}}}
& \rot{\textsc{\textbf{ev-last}}}
& \rot{\textsc{\textbf{shuffled}}}
& \rot{\textbf{Text}}
& \rot{\textbf{Visual}}
& \rot{\textbf{Layered}} \\

\midrule

Qwen2.5-VL-3B
& \cellcolor{lvqaBase!30}\stk{\textbf{.408}}{.02}
& \cellcolor{lvqaDown!11}\stk{.334}{.01}
& \cellcolor{lvqaDown!38}\stk{.142}{.01}
& \cellcolor{lvqaUp!12}\stk{.439}{.01}
& \cellcolor{lvqaDown!28}\stk{.213}{.05}
& \cellcolor{lvqaDown!23}\stk{.250}{.02}
& \cellcolor{lvqaDown!37}\stk{.150}{.02}
& \cellcolor{lvqaDown!5}\stk{.373}{.01}
& \cellcolor{lvqaDown!26}\stk{.224}{.03}
& \cellcolor{lvqaDown!30}\stk{.201}{.03}
& \cellcolor{lvqaDown!34}\stk{.170}{.01}
& \cellcolor{lvqaDown!45}\stk{.097}{.02}
& \cellcolor{lvqaDown!31}\stk{.193}{.03}
& \cellcolor{lvqaDown!11}\stk{.333}{.03}
& \cellcolor{lvqaDown!27}69
& \cellcolor{lvqaDown!34}61
& \cellcolor{lvqaDown!51}41 \\

Ministral-3-8B
& \cellcolor{lvqaBase!30}\stk{\textbf{.463}}{.03}
& \cellcolor{lvqaDown!4}\stk{.433}{.02}
& \cellcolor{lvqaDown!38}\stk{.199}{.04}
& \cellcolor{lvqaDown!9}\stk{.402}{.01}
& \cellcolor{lvqaDown!38}\stk{.200}{.03}
& \cellcolor{lvqaDown!31}\stk{.247}{.02}
& \cellcolor{lvqaDown!41}\stk{.180}{.01}
& \cellcolor{lvqaUp!23}\stk{.523}{.03}
& \cellcolor{lvqaDown!28}\stk{.272}{.01}
& \cellcolor{lvqaDown!46}\stk{.143}{.02}
& \cellcolor{lvqaDown!41}\stk{.176}{.01}
& \cellcolor{lvqaDown!50}\stk{.113}{.02}
& \cellcolor{lvqaDown!24}\stk{.298}{.02}
& \cellcolor{lvqaDown!1}\stk{.454}{.03}
& \cellcolor{lvqaDown!29}67
& \cellcolor{lvqaDown!29}66
& \cellcolor{lvqaDown!52}39 \\

Qwen2.5-VL-7B
& \cellcolor{lvqaBase!30}\stk{\textbf{.556}}{.02}
& \cellcolor{lvqaDown!9}\stk{.496}{.02}
& \cellcolor{lvqaDown!17}\stk{.439}{.03}
& \cellcolor{lvqaUp!45}\stk{.672}{.01}
& \cellcolor{lvqaDown!8}\stk{.497}{.01}
& \cellcolor{lvqaDown!31}\stk{.340}{.02}
& \cellcolor{lvqaDown!50}\stk{.208}{.03}
& \cellcolor{lvqaUp!36}\stk{.649}{.03}
& \cellcolor{lvqaDown!40}\stk{.280}{.01}
& \cellcolor{lvqaDown!36}\stk{.309}{.01}
& \cellcolor{lvqaDown!34}\stk{.321}{.03}
& \cellcolor{lvqaDown!64}\stk{.114}{.01}
& \cellcolor{lvqaDown!33}\stk{.329}{.01}
& \cellcolor{lvqaDown!11}\stk{.478}{.04}
& \cellcolor{lvqaDown!5}95
& \cellcolor{lvqaDown!29}66
& \cellcolor{lvqaDown!45}48 \\

Qwen2.5-VL-32B
& \cellcolor{lvqaBase!30}\stk{\textbf{.559}}{.01}
& \cellcolor{lvqaDown!1}\stk{.549}{.03}
& \cellcolor{lvqaDown!4}\stk{.533}{.00}
& \cellcolor{lvqaUp!58}\stk{.711}{.04}
& \cellcolor{lvqaUp!27}\stk{.630}{.03}
& \cellcolor{lvqaDown!19}\stk{.426}{.00}
& \cellcolor{lvqaDown!33}\stk{.327}{.02}
& \cellcolor{lvqaUp!58}\stk{.708}{.03}
& \cellcolor{lvqaDown!18}\stk{.433}{.00}
& \cellcolor{lvqaDown!32}\stk{.337}{.02}
& \cellcolor{lvqaDown!28}\stk{.367}{.02}
& \cellcolor{lvqaDown!41}\stk{.274}{.04}
& \cellcolor{lvqaDown!18}\stk{.432}{.02}
& \cellcolor{lvqaDown!0}\stk{.557}{.02}
& \cellcolor{lvqaUp!19}108
& \cellcolor{lvqaDown!13}85
& \cellcolor{lvqaDown!32}63 \\

Pixtral-12B
& \cellcolor{lvqaBase!30}\stk{\textbf{.572}}{.01}
& \cellcolor{lvqaDown!2}\stk{.556}{.01}
& \cellcolor{lvqaDown!26}\stk{.393}{.02}
& \cellcolor{lvqaUp!2}\stk{.577}{.02}
& \cellcolor{lvqaDown!15}\stk{.471}{.04}
& \cellcolor{lvqaDown!31}\stk{.360}{.01}
& \cellcolor{lvqaDown!49}\stk{.233}{.02}
& \cellcolor{lvqaUp!3}\stk{.579}{.01}
& \cellcolor{lvqaDown!26}\stk{.392}{.03}
& \cellcolor{lvqaDown!43}\stk{.271}{.01}
& \cellcolor{lvqaDown!39}\stk{.299}{.00}
& \cellcolor{lvqaDown!60}\stk{.154}{.01}
& \cellcolor{lvqaDown!43}\stk{.271}{.02}
& \cellcolor{lvqaDown!10}\stk{.502}{.01}
& \cellcolor{lvqaDown!11}87
& \cellcolor{lvqaDown!27}68
& \cellcolor{lvqaDown!49}43 \\

InternVL3.5-8B
& \cellcolor{lvqaBase!30}\stk{\textbf{.577}}{.01}
& \cellcolor{lvqaDown!8}\stk{.524}{.01}
& \cellcolor{lvqaDown!34}\stk{.343}{.01}
& \cellcolor{lvqaUp!46}\stk{.696}{.03}
& \cellcolor{lvqaDown!29}\stk{.376}{.00}
& \cellcolor{lvqaDown!28}\stk{.386}{.01}
& \cellcolor{lvqaDown!40}\stk{.300}{.02}
& \cellcolor{lvqaUp!16}\stk{.618}{.03}
& \cellcolor{lvqaDown!33}\stk{.349}{.05}
& \cellcolor{lvqaDown!24}\stk{.410}{.02}
& \cellcolor{lvqaDown!22}\stk{.424}{.03}
& \cellcolor{lvqaDown!42}\stk{.284}{.03}
& \cellcolor{lvqaDown!31}\stk{.362}{.03}
& \cellcolor{lvqaDown!14}\stk{.481}{.02}
& \cellcolor{lvqaDown!14}84
& \cellcolor{lvqaDown!25}72
& \cellcolor{lvqaDown!31}64 \\

Qwen3.5-9B
& \cellcolor{lvqaBase!30}\stk{\textbf{.656}}{.01}
& \cellcolor{lvqaDown!6}\stk{.612}{.00}
& \cellcolor{lvqaDown!16}\stk{.542}{.02}
& \cellcolor{lvqaUp!41}\stk{.762}{.01}
& \cellcolor{lvqaDown!3}\stk{.638}{.05}
& \cellcolor{lvqaDown!31}\stk{.443}{.02}
& \cellcolor{lvqaDown!37}\stk{.398}{.03}
& \cellcolor{lvqaUp!37}\stk{.751}{.05}
& \cellcolor{lvqaDown!24}\stk{.492}{.02}
& \cellcolor{lvqaDown!47}\stk{.332}{.02}
& \cellcolor{lvqaDown!47}\stk{.330}{.03}
& \cellcolor{lvqaDown!47}\stk{.329}{.02}
& \cellcolor{lvqaDown!22}\stk{.504}{.02}
& \cellcolor{lvqaDown!7}\stk{.607}{.01}
& \cellcolor{lvqaDown!2}97
& \cellcolor{lvqaDown!18}79
& \cellcolor{lvqaDown!37}57 \\

GPT-5.6-Luna$^{\dagger}$
& \cellcolor{lvqaBase!30}\stk{\textbf{.693}}{.01}
& \cellcolor{lvqaUp!0}\stk{.694}{.01}
& \cellcolor{lvqaDown!6}\stk{.654}{.01}
& \cellcolor{lvqaUp!8}\stk{.715}{.01}
& \cellcolor{lvqaDown!2}\stk{.681}{.02}
& \cellcolor{lvqaDown!7}\stk{.642}{.01}
& \cellcolor{lvqaDown!10}\stk{.621}{.02}
& \cellcolor{lvqaDown!1}\stk{.689}{.01}
& \cellcolor{lvqaDown!8}\stk{.638}{.01}
& \cellcolor{lvqaDown!19}\stk{.563}{.02}
& \cellcolor{lvqaDown!16}\stk{.585}{.03}
& \cellcolor{lvqaDown!10}\stk{.621}{.02}
& \cellcolor{lvqaDown!11}\stk{.619}{.02}
& \cellcolor{lvqaDown!1}\stk{.688}{.03}
& \cellcolor{lvqaDown!1}99
& \cellcolor{lvqaDown!6}93
& \cellcolor{lvqaDown!12}86 \\

Qwen3.6-27B
& \cellcolor{lvqaBase!30}\stk{\textbf{.697}}{.02}
& \cellcolor{lvqaDown!3}\stk{.679}{.02}
& \cellcolor{lvqaDown!24}\stk{.533}{.01}
& \cellcolor{lvqaUp!34}\stk{.783}{.02}
& \cellcolor{lvqaDown!7}\stk{.650}{.02}
& \cellcolor{lvqaDown!28}\stk{.502}{.01}
& \cellcolor{lvqaDown!34}\stk{.460}{.01}
& \cellcolor{lvqaUp!25}\stk{.762}{.02}
& \cellcolor{lvqaDown!19}\stk{.563}{.01}
& \cellcolor{lvqaDown!47}\stk{.370}{.03}
& \cellcolor{lvqaDown!44}\stk{.391}{.02}
& \cellcolor{lvqaDown!40}\stk{.417}{.02}
& \cellcolor{lvqaDown!24}\stk{.529}{.01}
& \cellcolor{lvqaDown!2}\stk{.683}{.01}
& \cellcolor{lvqaDown!4}95
& \cellcolor{lvqaDown!15}82
& \cellcolor{lvqaDown!33}61 \\

Qwen3-VL-4B
& \cellcolor{lvqaBase!30}\stk{\textbf{.703}}{.01}
& \cellcolor{lvqaDown!7}\stk{.656}{.00}
& \cellcolor{lvqaDown!12}\stk{.618}{.02}
& \cellcolor{lvqaUp!24}\stk{.765}{.01}
& \cellcolor{lvqaUp!14}\stk{.739}{.03}
& \cellcolor{lvqaDown!27}\stk{.513}{.01}
& \cellcolor{lvqaDown!57}\stk{.308}{.00}
& \cellcolor{lvqaUp!30}\stk{.781}{.00}
& \cellcolor{lvqaDown!27}\stk{.513}{.04}
& \cellcolor{lvqaDown!49}\stk{.362}{.04}
& \cellcolor{lvqaDown!41}\stk{.420}{.02}
& \cellcolor{lvqaDown!71}\stk{.212}{.01}
& \cellcolor{lvqaDown!25}\stk{.528}{.01}
& \cellcolor{lvqaDown!7}\stk{.657}{.00}
& \cellcolor{lvqaDown!1}99
& \cellcolor{lvqaDown!21}75
& \cellcolor{lvqaDown!40}54 \\

Qwen3-VL-30B
& \cellcolor{lvqaBase!30}\stk{\textbf{.722}}{.00}
& \cellcolor{lvqaDown!7}\stk{.676}{.01}
& \cellcolor{lvqaDown!18}\stk{.600}{.01}
& \cellcolor{lvqaUp!15}\stk{.760}{.02}
& \cellcolor{lvqaDown!1}\stk{.716}{.02}
& \cellcolor{lvqaDown!22}\stk{.567}{.01}
& \cellcolor{lvqaDown!36}\stk{.476}{.01}
& \cellcolor{lvqaUp!8}\stk{.742}{.02}
& \cellcolor{lvqaDown!32}\stk{.501}{.01}
& \cellcolor{lvqaDown!30}\stk{.511}{.02}
& \cellcolor{lvqaDown!24}\stk{.556}{.02}
& \cellcolor{lvqaDown!35}\stk{.479}{.03}
& \cellcolor{lvqaDown!32}\stk{.501}{.03}
& \cellcolor{lvqaDown!8}\stk{.664}{.01}
& \cellcolor{lvqaDown!4}95
& \cellcolor{lvqaDown!18}79
& \cellcolor{lvqaDown!25}71 \\

Qwen3-VL-8B
& \cellcolor{lvqaBase!30}\stk{\textbf{.729}}{.01}
& \cellcolor{lvqaDown!7}\stk{.681}{.00}
& \cellcolor{lvqaDown!1}\stk{.721}{.01}
& \cellcolor{lvqaUp!21}\stk{.782}{.03}
& \cellcolor{lvqaDown!4}\stk{.702}{.03}
& \cellcolor{lvqaDown!26}\stk{.550}{.01}
& \cellcolor{lvqaDown!43}\stk{.433}{.01}
& \cellcolor{lvqaUp!14}\stk{.764}{.02}
& \cellcolor{lvqaDown!26}\stk{.550}{.00}
& \cellcolor{lvqaDown!28}\stk{.538}{.02}
& \cellcolor{lvqaDown!24}\stk{.564}{.02}
& \cellcolor{lvqaDown!41}\stk{.444}{.03}
& \cellcolor{lvqaDown!24}\stk{.560}{.03}
& \cellcolor{lvqaDown!5}\stk{.694}{.00}
& \cellcolor{lvqaDown!1}99
& \cellcolor{lvqaDown!18}79
& \cellcolor{lvqaDown!24}72 \\

Gemini-3.7-Flash$^{\dagger}$
& \cellcolor{lvqaBase!30}\stk{\textbf{.862}}{.01}
& \cellcolor{lvqaDown!0}\stk{.858}{.01}
& \cellcolor{lvqaDown!4}\stk{.835}{.00}
& \cellcolor{lvqaDown!0}\stk{.860}{.00}
& \cellcolor{lvqaDown!2}\stk{.844}{.01}
& \cellcolor{lvqaDown!2}\stk{.850}{.01}
& \cellcolor{lvqaDown!4}\stk{.836}{.00}
& \cellcolor{lvqaDown!0}\stk{.860}{.01}
& \cellcolor{lvqaDown!4}\stk{.835}{.00}
& \cellcolor{lvqaDown!8}\stk{.805}{.01}
& \cellcolor{lvqaDown!11}\stk{.783}{.00}
& \cellcolor{lvqaDown!3}\stk{.842}{.01}
& \cellcolor{lvqaDown!2}\stk{.847}{.01}
& \cellcolor{lvqaUp!0}\stk{.862}{.01}
& \cellcolor{lvqaDown!1}99
& \cellcolor{lvqaDown!2}98
& \cellcolor{lvqaDown!4}95 \\

\midrule

\textit{Mean, Open}
& \cellcolor{lvqaBase!30}\textbf{.604}
& \cellcolor{lvqaDown!5}.563
& \cellcolor{lvqaDown!18}.460
& \cellcolor{lvqaUp!22}.668
& \cellcolor{lvqaDown!9}.530
& \cellcolor{lvqaDown!23}.417
& \cellcolor{lvqaDown!36}.316
& \cellcolor{lvqaUp!18}.659
& \cellcolor{lvqaDown!24}.415
& \cellcolor{lvqaDown!34}.344
& \cellcolor{lvqaDown!31}.365
& \cellcolor{lvqaDown!42}.265
& \cellcolor{lvqaDown!25}.410
& \cellcolor{lvqaDown!6}.555
& \cellcolor{lvqaDown!7}90
& \cellcolor{lvqaDown!20}74
& \cellcolor{lvqaDown!34}56 \\

\textit{$\Delta$ vs \textsc{full}}
& ---
& -.041
& -.144
& +.064
& -.074
& -.187
& -.288
& +.055
& -.189
& -.260
& -.239
& -.339
& -.194
& -.049
& {}
& {}
& {} \\

\bottomrule
\end{tabular}%
}
\vspace{-5mm}
\end{table*}

%% file: tables/Exp-A.tex
\begin{wraptable}{r}{0.62\textwidth}
\centering
\small
\vspace{-5mm}
\caption{\textbf{What models can do when evidence is handed to them.}}
\label{tab:exp-a}
\renewcommand{\arraystretch}{0.95}
\resizebox{\linewidth}{!}{%
\begin{tabular}{@{}l*{12}{Z}@{}}
\toprule
&
\multicolumn{3}{c}{\grp{\textbf{answer accuracy}}}
&
\multicolumn{5}{c}{\grp{\textbf{evidence selection}}}
&
\multicolumn{3}{c}{\grp{\textbf{error modes}}}
&
\multicolumn{1}{c}{\grp{\textbf{joint}}}
\\

\cmidrule(lr){2-4}
\cmidrule(lr){5-9}
\cmidrule(lr){10-12}
\cmidrule(l){13-13}

\textbf{Model}
& \multicolumn{1}{c}{\rot{\textsc{\textbf{rgb}}}}
& \multicolumn{1}{c}{\rot{\textsc{\textbf{all-lyr}}}}
& \multicolumn{1}{c}{\rot{\textsc{\textbf{oracle}}}}
& \multicolumn{1}{c}{\rot{\textbf{Precision}}}
& \multicolumn{1}{c}{\rot{\textbf{Recall}}}
& \multicolumn{1}{c}{\rot{\textbf{F1-score}}}
& \multicolumn{1}{c}{\rot{\textbf{Ex Match}}}
& \multicolumn{1}{c}{\rot{\textbf{Comp.}}}
& \multicolumn{1}{c}{\rot{\textbf{Exc.}}}
& \multicolumn{1}{c}{\rot{\textbf{Dist.}}}
& \multicolumn{1}{c}{\rot{\textbf{Inv.}}}
& \multicolumn{1}{c}{\rot{\textbf{Acc.}}}
\\

\midrule

Q2.5-VL-3B
& \cellcolor{lvqaBase!30}\stk{\textbf{.408}}{.02}
& \cellcolor{lvqaDown!5}\stk{.371}{.03}
& \cellcolor{lvqaDown!15}\stk{.303}{.01}
& \cellcolor{lvqaDown!12}\stk{.561}{.01}
& \cellcolor{lvqaUp!0}\stk{.522}{.01}
& \cellcolor{lvqaDown!7}\stk{.503}{.01}
& \cellcolor{lvqaDown!17}\stk{.101}{.01}
& \cellcolor{lvqaDown!1}\stk{.279}{.00}
& \cellcolor{lvqaDown!8}\stk{.387}{.02}
& \cellcolor{lvqaDown!38}\stk{.520}{.02}
& \cellcolor{lvqaUp!15}\stk{.137}{.01}
& \cellcolor{lvqaDown!16}\stk{.048}{.01}
\\

Min-3-8B
& \cellcolor{lvqaBase!30}\stk{\textbf{.463}}{.03}
& \cellcolor{lvqaDown!6}\stk{.420}{.02}
& \cellcolor{lvqaDown!28}\stk{.267}{.02}
& \cellcolor{lvqaDown!2}\stk{.617}{.01}
& \cellcolor{lvqaUp!9}\stk{.565}{.01}
& \cellcolor{lvqaUp!4}\stk{.562}{.01}
& \cellcolor{lvqaDown!4}\stk{.174}{.00}
& \cellcolor{lvqaUp!3}\stk{.295}{.01}
& \cellcolor{lvqaUp!5}\stk{.318}{.01}
& \cellcolor{lvqaDown!15}\stk{.409}{.02}
& \cellcolor{lvqaUp!11}\stk{.159}{.02}
& \cellcolor{lvqaDown!11}\stk{.080}{.01}
\\

Q2.5-VL-7B
& \cellcolor{lvqaBase!30}\stk{\textbf{.556}}{.02}
& \cellcolor{lvqaDown!13}\stk{.463}{.02}
& \cellcolor{lvqaDown!26}\stk{.372}{.01}
& \cellcolor{lvqaDown!28}\stk{.469}{.01}
& \cellcolor{lvqaDown!38}\stk{.364}{.02}
& \cellcolor{lvqaDown!32}\stk{.385}{.02}
& \cellcolor{lvqaDown!17}\stk{.105}{.01}
& \cellcolor{lvqaDown!29}\stk{.154}{.02}
& \cellcolor{lvqaDown!9}\stk{.395}{.02}
& \cellcolor{lvqaUp!1}\stk{.327}{.02}
& \cellcolor{lvqaDown!6}\stk{.251}{.01}
& \cellcolor{lvqaDown!14}\stk{.059}{.01}
\\

Q2.5-VL-32B
& \cellcolor{lvqaBase!30}\stk{\textbf{.559}}{.01}
& \cellcolor{lvqaUp!27}\stk{.629}{.01}
& \cellcolor{lvqaDown!12}\stk{.474}{.02}
& \cellcolor{lvqaDown!19}\stk{.519}{.01}
& \cellcolor{lvqaDown!20}\stk{.437}{.02}
& \cellcolor{lvqaDown!19}\stk{.447}{.01}
& \cellcolor{lvqaDown!13}\stk{.126}{.02}
& \cellcolor{lvqaDown!17}\stk{.205}{.03}
& \cellcolor{lvqaDown!20}\stk{.455}{.01}
& \cellcolor{lvqaUp!4}\stk{.314}{.02}
& \cellcolor{lvqaDown!38}\stk{.413}{.03}
& \cellcolor{lvqaDown!10}\stk{.083}{.02}
\\

Pixtral-12B
& \cellcolor{lvqaBase!30}\stk{\textbf{.572}}{.01}
& \cellcolor{lvqaDown!14}\stk{.478}{.02}
& \cellcolor{lvqaDown!25}\stk{.396}{.03}
& \cellcolor{lvqaUp!3}\stk{.649}{.03}
& \cellcolor{lvqaDown!7}\stk{.494}{.01}
& \cellcolor{lvqaDown!2}\stk{.531}{.02}
& \cellcolor{lvqaDown!5}\stk{.168}{.02}
& \cellcolor{lvqaDown!12}\stk{.230}{.01}
& \cellcolor{lvqaUp!4}\stk{.324}{.03}
& \cellcolor{lvqaDown!10}\stk{.383}{.02}
& \cellcolor{lvqaUp!23}\stk{.092}{.02}
& \cellcolor{lvqaDown!10}\stk{.087}{.02}
\\

IVL3.5-8B
& \cellcolor{lvqaBase!30}\stk{\textbf{.577}}{.01}
& \cellcolor{lvqaDown!1}\stk{.568}{.00}
& \cellcolor{lvqaDown!24}\stk{.410}{.00}
& \cellcolor{lvqaDown!11}\stk{.563}{.02}
& \cellcolor{lvqaDown!38}\stk{.365}{.01}
& \cellcolor{lvqaDown!26}\stk{.416}{.01}
& \cellcolor{lvqaDown!17}\stk{.102}{.02}
& \cellcolor{lvqaDown!33}\stk{.134}{.02}
& \cellcolor{lvqaDown!16}\stk{.433}{.02}
& \cellcolor{lvqaUp!0}\stk{.333}{.02}
& \cellcolor{lvqaDown!6}\stk{.254}{.02}
& \cellcolor{lvqaDown!15}\stk{.056}{.02}
\\

Q3.5-9B
& \cellcolor{lvqaBase!30}\stk{\textbf{.656}}{.01}
& \cellcolor{lvqaDown!3}\stk{.634}{.02}
& \cellcolor{lvqaDown!24}\stk{.489}{.03}
& \cellcolor{lvqaDown!3}\stk{.609}{.02}
& \cellcolor{lvqaUp!2}\stk{.531}{.01}
& \cellcolor{lvqaDown!1}\stk{.533}{.01}
& \cellcolor{lvqaDown!12}\stk{.129}{.01}
& \cellcolor{lvqaDown!3}\stk{.270}{.01}
& \cellcolor{lvqaDown!5}\stk{.375}{.01}
& \cellcolor{lvqaDown!18}\stk{.421}{.03}
& \cellcolor{lvqaDown!1}\stk{.229}{.00}
& \cellcolor{lvqaDown!10}\stk{.085}{.01}
\\

GPT-5.6-L$^{\dagger}$
& \cellcolor{lvqaBase!30}\stk{\textbf{.693}}{.01}
& \cellcolor{lvqaUp!10}\stk{.719}{.01}
& \cellcolor{lvqaDown!8}\stk{.634}{.01}
& \cellcolor{lvqaUp!45}\stk{.927}{.01}
& \cellcolor{lvqaUp!44}\stk{.729}{.02}
& \cellcolor{lvqaUp!45}\stk{.785}{.02}
& \cellcolor{lvqaUp!43}\stk{.468}{.02}
& \cellcolor{lvqaUp!41}\stk{.491}{.02}
& \cellcolor{lvqaUp!44}\stk{.073}{.01}
& \cellcolor{lvqaUp!38}\stk{.122}{.01}
& \cellcolor{lvqaUp!39}\stk{.000}{.00}
& \cellcolor{lvqaUp!32}\stk{.348}{.01}
\\

Q3.6-27B
& \cellcolor{lvqaBase!30}\stk{\textbf{.697}}{.02}
& \cellcolor{lvqaUp!15}\stk{.736}{.02}
& \cellcolor{lvqaDown!23}\stk{.539}{.02}
& \cellcolor{lvqaUp!14}\stk{.722}{.01}
& \cellcolor{lvqaUp!29}\stk{.658}{.00}
& \cellcolor{lvqaUp!21}\stk{.654}{.00}
& \cellcolor{lvqaUp!8}\stk{.247}{.01}
& \cellcolor{lvqaUp!24}\stk{.406}{.01}
& \cellcolor{lvqaUp!12}\stk{.275}{.01}
& \cellcolor{lvqaDown!14}\stk{.404}{.02}
& \cellcolor{lvqaUp!19}\stk{.116}{.01}
& \cellcolor{lvqaUp!8}\stk{.191}{.00}
\\

Q3-VL-4B
& \cellcolor{lvqaBase!30}\stk{\textbf{.703}}{.01}
& \cellcolor{lvqaDown!1}\stk{.693}{.01}
& \cellcolor{lvqaDown!19}\stk{.572}{.01}
& \cellcolor{lvqaDown!35}\stk{.424}{.00}
& \cellcolor{lvqaDown!49}\stk{.317}{.00}
& \cellcolor{lvqaDown!40}\stk{.346}{.00}
& \cellcolor{lvqaDown!19}\stk{.091}{.01}
& \cellcolor{lvqaDown!38}\stk{.113}{.00}
& \cellcolor{lvqaDown!42}\stk{.569}{.00}
& \cellcolor{lvqaUp!10}\stk{.279}{.02}
& \cellcolor{lvqaDown!60}\stk{.522}{.01}
& \cellcolor{lvqaDown!14}\stk{.060}{.01}
\\

Q3-VL-30B
& \cellcolor{lvqaBase!30}\stk{\textbf{.722}}{.00}
& \cellcolor{lvqaDown!1}\stk{.717}{.01}
& \cellcolor{lvqaDown!17}\stk{.607}{.02}
& \cellcolor{lvqaDown!11}\stk{.562}{.00}
& \cellcolor{lvqaDown!1}\stk{.517}{.01}
& \cellcolor{lvqaDown!6}\stk{.510}{.00}
& \cellcolor{lvqaDown!7}\stk{.160}{.02}
& \cellcolor{lvqaDown!0}\stk{.281}{.02}
& \cellcolor{lvqaDown!17}\stk{.438}{.00}
& \cellcolor{lvqaDown!8}\stk{.371}{.01}
& \cellcolor{lvqaDown!40}\stk{.424}{.01}
& \cellcolor{lvqaDown!4}\stk{.120}{.01}
\\

Q3-VL-8B
& \cellcolor{lvqaBase!30}\stk{\textbf{.729}}{.01}
& \cellcolor{lvqaUp!6}\stk{.744}{.03}
& \cellcolor{lvqaDown!21}\stk{.580}{.01}
& \cellcolor{lvqaDown!11}\stk{.563}{.01}
& \cellcolor{lvqaDown!5}\stk{.501}{.01}
& \cellcolor{lvqaDown!7}\stk{.505}{.01}
& \cellcolor{lvqaDown!9}\stk{.149}{.01}
& \cellcolor{lvqaDown!6}\stk{.254}{.01}
& \cellcolor{lvqaDown!16}\stk{.430}{.01}
& \cellcolor{lvqaDown!17}\stk{.414}{.02}
& \cellcolor{lvqaDown!15}\stk{.298}{.00}
& \cellcolor{lvqaDown!3}\stk{.124}{.02}
\\

Gem-3.7-F$^{\dagger}$
& \cellcolor{lvqaBase!30}\stk{\textbf{.862}}{.01}
& \cellcolor{lvqaUp!8}\stk{.883}{.00}
& \cellcolor{lvqaDown!4}\stk{.833}{.01}
& \cellcolor{lvqaUp!52}\stk{.971}{.00}
& \cellcolor{lvqaUp!52}\stk{.769}{.01}
& \cellcolor{lvqaUp!52}\stk{.826}{.01}
& \cellcolor{lvqaUp!52}\stk{.521}{.01}
& \cellcolor{lvqaUp!52}\stk{.549}{.01}
& \cellcolor{lvqaUp!52}\stk{.024}{.00}
& \cellcolor{lvqaUp!52}\stk{.043}{.00}
& \cellcolor{lvqaUp!39}\stk{.000}{.00}
& \cellcolor{lvqaUp!52}\stk{.474}{.01}
\\

\midrule

\textit{Mean, Open}
& \cellcolor{lvqaBase!30}\textbf{.604}
& \cellcolor{lvqaDown!2}.587
& \cellcolor{lvqaDown!19}.455
& .569
& .479
& .490
& .141
& .238
& .400
& .380
& .263
& .090
\\

\bottomrule

\end{tabular}%
}

\vspace{-3mm}
\end{wraptable}

%% file: sections/6-conclusion.tex
\vspace{-10pt}
\section{Conclusion}
\label{sec:conclusion}

We built Layered-VQA to separate two things that multi-turn evaluation normally changes at once.
Composition is whether the model sees the scene whole or in pieces.
Conversation is whether those pieces arrive in one turn or across many.
Sixteen delivery conditions hold the question content and the source pixels fixed, which isolates the effect of presentation more closely than standard multi-turn benchmarks.
Conversation is cheaper: splitting the question costs $-.041$, while splitting the scene costs $-.187$ even when all layers arrive in one turn.
Recomposing the scene recovers $+.343$ without adding new pixels: the loss comes from fragmentation, not missing evidence.
Two consequences follow, and neither is visible to a benchmark that shows one image and one question.
Even oracle-selected sufficient layers trail the whole scene by $-.148$, showing that correct evidence selection can still hurt performance.
Separately, a right answer is not a grounded one, since nearly three in ten correct answers arrive before the evidence supporting them does.
Both failures concentrate on questions about relations between objects, which lose roughly three times more than questions about a single object.
Fragmenting a scene keeps the objects and destroys what holds between them.
VLMs are increasingly deployed on evidence that is cropped, retrieved, or revealed over time.
Our results say that pipeline design carries a cost that answer accuracy alone will not report.
Preserve the composed view, recompose before asking, and score grounding beside accuracy.
\emph{\textbf{The right evidence is not enough: VLMs need the scene it came from.}}

%% file: sections/ethics-reproduce-ai.tex
% \section*{Limitations}
% \label{sec:limitations}

% The layers come from one annotated source, so a scene is decomposed the way that annotation decomposes it.
% A different segmentation would draw different boundaries and could shift how much a single layer carries.
% The proprietary models are assistants only.
% The following two conditions are not fully controlled.
% \textsc{misaligned} front-loads the gold layers relative to \textsc{aligned}, so the two are not matched on when the evidence arrives and we make no claim from that pair.
% The recap conditions restore the scene only at the last turn, so we measure that the loss is recoverable and not how quickly it becomes unrecoverable.
% Finally, the benchmark asks about static scenes with a fixed layer inventory.
% Whether the same compositional cost appears when a model can choose what to look at next is an open question, and one the design here was not built to answer.

\section*{Ethics Statement}

The scenes come from \textsc{LAION} and \textsc{LVIS} through the \textsc{MuLAn} layer decomposition, all publicly released datasets.
Many contain people, photographed in public settings by the original collectors.
We add no new imagery, collect no personal data, and identify no one.
The questions and reference answers were written by the authors, and the layer annotations were produced and checked by the authors.
No crowd workers or external annotators were employed, so there are no labour concerns to report.
The benchmark measures a failure mode rather than proposing a deployment.
One result invites misreading and we state it plainly: an oracle that returns exactly the right evidence performs worse than the whole scene.
That is an argument for keeping context, not an argument against retrieval.
The proprietary models were accessed through a commercial API under its terms of service, and we report their behaviour without claiming access to their weights or training data.

\section*{Reproducibility Statement}

The supplementary material contains the full code and the benchmark itself.
The $16$ delivery conditions are shipped as JSON, one file per condition, with the question shards, the layer schedules, the gold and minimal sufficient layer sets, and the distractor layers.
Everything is included: the simulator, the user agents, the judges, the scoring, the annotation pipeline, and the scripts that produce every table and figure in this paper from the raw logs.
The settings needed to repeat a run are in Section~\ref{sec:experimental_setup}, and every prompt is reproduced in Appendix~\ref{app:prompts}.
Each model runs three times under identical settings, and we report the run-to-run spread beside every number.
Confidence intervals resample scenes rather than questions, since questions on one scene share an image.
We will release the code, the benchmark, and the images publicly with the paper.

\section*{AI Use Statement}

We used an AI assistant while writing the code and while drafting parts of the text.
The research itself was done by the authors: the idea, the experimental design, what each condition isolates, the analyses, the claims, and what the paper argues.
The assistant wrote code alongside us and helped tighten prose after the structure and the content of a section were decided.
Every line of code and every sentence in this paper was read and approved by the authors before it was included.
We take full responsibility for the paper, including anything the assistant contributed.
Vision-language models are also the object of study here rather than a tool.
Their use as assistants, as simulated users, and as judges is described in Section~\ref{sec:experimental_setup}.

%% file: sections/appendix-additional-0.tex
\section{Anatomy of the loss}
\label{app:tables}

\subsection{Where the loss falls}

\input{tables/category}
Table~\ref{tab:category} splits every headline condition by question type.
It is the reason the paper claims a mechanism rather than a list of drops.
Questions that ask about one object on its own lose $-.081$ under \textsc{oracle} and $-.126$ under \textsc{concat-v}.
Questions that ask about a relation between objects lose $-.228$ and $-.259$.
The two losses fall on the same questions, in the same order, and they are roughly three times larger on relational questions than on single-object ones.
An oracle crop and a layer decomposition are both ways of fragmenting a scene, and fragmenting a scene keeps the objects and destroys what holds between them.
Rescheduling the question does not do this, which is why \textsc{concat-q} costs about $-.03$ to $-.05$ regardless of type.

\subsection{Response length across attempts}
\label{app:bloat}

\begin{figure}[h]
\centering
\includegraphics[width=\linewidth]{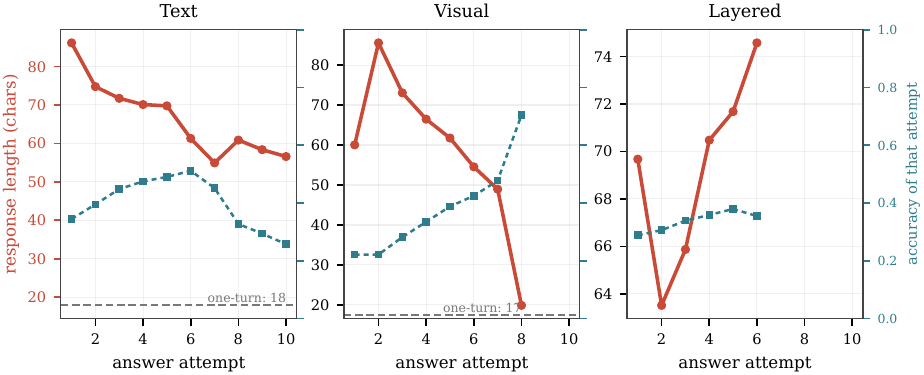}
\caption{\textbf{Response length and accuracy across successive answer attempts.}
The dashed line is the single-turn baseline for the same questions.
Attempts get terser, and accuracy peaks mid-conversation.}
\label{fig:bloat}
\end{figure}

Figure~\ref{fig:bloat} plots the length of each successive answer attempt against how often that attempt was correct.
Lost-in-Conversation~\cite{laban2026llms} reports that answers grow as a conversation drags on.
Ours shrink.
In the text conditions the first attempt averages $86$ characters and the tenth averages $57$.
The visual and layered conditions do the same.
The bloat appears immediately.
The same question answered in a single turn gets $18$ characters.
Every multi-turn first attempt is three to five times longer than that.
A model given part of a question pads immediately rather than gradually.
Accuracy of an attempt peaks mid-conversation.
In the text conditions it climbs from $.34$ on the first attempt to $.51$ on the sixth, then falls to $.26$ by the tenth.
A model still revising after six attempts is not converging on an answer, it is drifting away from one.

\subsection{Aptitude and unreliability}
\label{app:reliability}

\begin{figure}[h]
\centering
\includegraphics[width=\linewidth]{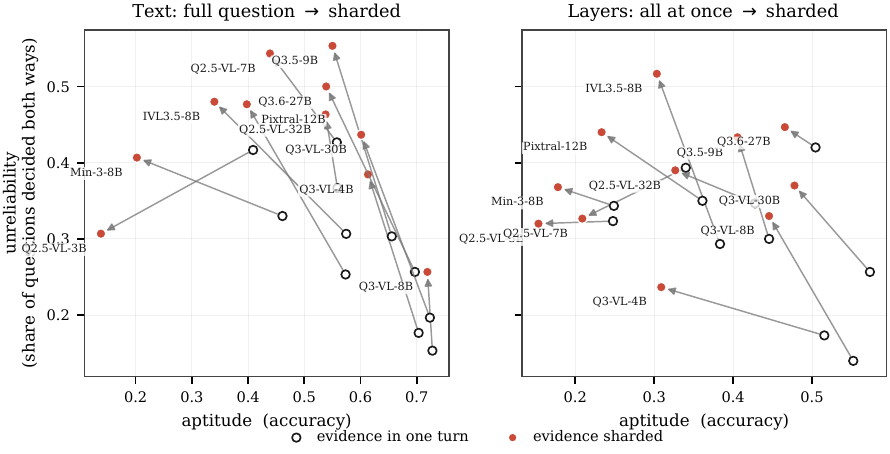}
\caption{\textbf{Aptitude against unreliability.}
Hollow markers deliver the evidence in one turn, filled markers shard it.
An arrow joins the two states of one model.}
\label{fig:reliability}
\end{figure}

Figure~\ref{fig:reliability} separates how well a model does from how consistently it does it.
Aptitude is accuracy.
Unreliability is the share of questions the model decides both ways across the three replicate runs, on the same question under identical settings.
Three exact replicates make this a direct measurement rather than a spread over resampled simulations.
Breaking the question into shards costs $.142$ of accuracy and adds $.147$ of unreliability, taking the field from $.290$ to $.437$.
Ten of the eleven models get less reliable.
Breaking the scene into layers costs $.099$ and adds $.076$, and nine of eleven get less reliable.
Almost every arrow points left and up.
No model trades accuracy for consistency.
Sharding does not make a model cautious, it makes it inconsistent.

\subsection{Premature commitment}
\label{app:premature}
Table~\ref{tab:first-answer-attempt} bins every conversation by how far through it the model first committed to an answer, then reports accuracy in each fifth.
\textbf{\textit{The best moment to answer is not the last one.}}
Accuracy runs $.388$, $.458$, $.460$, $.473$, $.403$.
It rises, peaks in the fourth fifth, and then falls again.
The text-only result this parallels is monotone: waiting longer is always better there.
Ours is not, and the difference is the point.
A text conversation withholds constraints, so a later answer is simply a better informed one.
Ours withholds a scene, and a scene has to be assembled.
A model that has still not committed at four fifths is usually one that could not assemble it, working from a transcript full of its own earlier hedging.
\textbf{\textit{Models commit before the evidence justifies it.}}
$62\%$ of first commitments land in the first two fifths, and the modal bin is the second fifth for ten of the eleven open models.
The first fifth is also the least accurate bin for nine of them.
Models are answering where they do worst, and they are doing it because our instructions ask every turn for either an answer or one clarifying question.
That is the same contract a deployed assistant operates under.
Nothing in the prompt says an answer is due, and they give one anyway.
\input{tables/first-attempt}
\textbf{\textit{What the ideal looks like.}}
A model should hold a hypothesis without stating it and commit when the last supporting layer lands, not before and not long after.
It should be able to say that it has seen a constraint and is still missing the evidence for it.
\textbf{\textit{What would get us there.}}
Make abstention cheap.
Instruction tuning rewards a helpful answer on every turn, which is exactly the behaviour that costs accuracy here.
A model that could emit an explicit \emph{evidence not yet sufficient} state, and not be penalised for it, would move its commitments out of the worst bin.
This bin is chosen by the model rather than assigned by us, so the table is observational.
The intervention that would make it causal is to force commitment at a fixed turn, which we leave to future work.

\subsection{Verbosity}
\label{app:verbosity}
Table~\ref{tab:verbosity} sorts each model's responses into length quintiles computed within that model, so a bin compares a model against itself rather than against a chattier one.
\textbf{\textit{Length is only bad once the scene is gone.}}
When the evidence arrives in one turn, accuracy runs $.558$, $.577$, $.589$, $.544$, $.408$.
It rises before it falls, and the median-length fifth is the best.
When the same evidence arrives across turns, it runs $.493$, $.484$, $.465$, $.447$, $.410$, falling from the first bin onward.
The prevailing advice from text-only work, that shorter responses are better, holds only in the second case.
In the first, some length is reasoning and it pays.
\textbf{\textit{Why the peak disappears.}}
Length in a single turn buys the model something to do: it can describe the scene, bind objects, and check a relation.
Length across turns buys nothing, because the scene it would reason over has not been assembled.
What remains is hedging, and hedging starts on the first reply rather than growing into it.
The drop from shortest to longest is negative for all eleven models, reaching $-.193$ across turns and $-.297$ within one.
\textbf{\textit{What the ideal looks like.}}
Response length should track how hard the question is, not how the evidence was delivered.
The same question should not produce a longer answer because the picture arrived in pieces.
\textbf{\textit{What would get us there.}}
Do not optimise length globally.
A brevity penalty applied everywhere would remove the reasoning that helps in the single-turn case along with the hedging that hurts in the multi-turn one.
Length is more useful as a signal than as a target: a reply that runs long under fragmentation, against the same model's own baseline on the same question, is a cheap runtime indicator that the model is under-evidenced and should be handed the composed scene.

\input{tables/verbosity}

\subsection{Ungrounded correctness}
\label{app:grounded}
Table~\ref{tab:ungrounded-grounded} asks, of the answers we score as correct, how many were already given before the layer supporting them had been delivered.
It covers only the seven conditions where the question is defined: multi-turn, and carrying a layer schedule.
\textbf{\textit{Answer accuracy is inflated by a prior.}}
Across those conditions accuracy is $.401$, and $.287$ once ungrounded answers are removed.
Nearly three in ten correct answers were produced before their evidence existed.
The model is not reading the image at that moment.
It is answering from what scenes of this kind usually contain, and our scenes are natural enough that this often works.
\textbf{\textit{Pairing evidence with the constraint that needs it helps.}}
The share is $.352$ where layers arrive on a fixed schedule unrelated to the question, $.282$ where the order of the layers is manipulated, and $.153$ where every turn carries both a piece of the question and a layer.
Only the last of these puts an image in front of the model at the moment a constraint is stated.
It does not raise accuracy much, and it changes what accuracy means.
\input{tables/ungrounded-correctness}
\textbf{\textit{The weakest models guess the most.}}
Ungrounded share tracks capability inversely, from $.421$ for the smallest model down to $.202$ for the strongest open one.
It does not vanish with scale.
Gemini-3.7-Flash answers correctly $.830$ of the time and grounded $.588$ of the time, the largest absolute grounding gap in the study, because a model that is right more often has more correct answers available to be ungrounded.
\textbf{\textit{What the ideal looks like.}}
Scored accuracy and grounded accuracy should coincide.
A model should decline to answer until the layer carrying the answer has arrived, and should be able to name it.
\textbf{\textit{What would get us there.}}
Report the gap.
It is a hallucination rate measured on a benchmark where the ground truth for \emph{which evidence was required} is annotated, which is rare, and it costs nothing beyond what a normal evaluation already computes.
Training against it is the harder half: a model rewarded only for the final answer has no reason to wait, since a lucky prior scores the same as a grounded reading.
A citation requirement of the kind \textsc{all-layers} imposes turns that into something optimisable.

\subsection{Self-preference among judges}
\label{app:judges}

In the primary runs each model grades its own conversations.
That is convenient and it is also a conflict of interest.
To measure it we re-graded every conversation with all eleven models and built the $11 \times 11$ matrix in Figure~\ref{fig:selfpref}.
A raw cell mixes three things: the assistant was good, the judge was generous, or the judge favoured that assistant.
Only the third is interesting, so we remove the other two.
For each cell we subtract the assistant's mean score across all judges and the judge's mean score across all assistants, then add the grand mean back.
What is left is the residual, and it is what the figure plots.
\textbf{\textit{Judges disagree more than models do.}}
Leniency runs from $.404$ to $.720$ across judges, a spread of $.32$.
Ability runs from $.334$ to $.629$ across assistants, a spread of $.30$.
Picking a judge therefore moves a score about as much as picking a model.
A benchmark graded by one judge is reporting that judge as much as the field.
\textbf{\textit{Every model favours itself.}}
All eleven diagonal residuals are positive.
The average model gains $2.6$ points from grading its own work, and off-diagonal residuals average zero.
Nothing else in the matrix has a sign that holds eleven times out of eleven.
\textbf{\textit{The bias is largest where it does the most damage.}}
Qwen2.5-VL-3B gains $12.9$ points on itself, the largest value in the matrix, and it is the weakest model in the study.
Ministral-3-8B gains $0.5$ points, the smallest.
Self-grading does not add noise evenly.
It lifts the bottom of the ranking, which is exactly where a benchmark needs to discriminate.
This is why every number for the open models is a consensus verdict rather than a self-graded one. The self-graded results are only for the closed-models.
A model still grades itself in one cell of eleven, and ten other judges outvote it.

\subsection{Human validation of the consensus judge}
\label{app:human}

Every open model accuracy is a consensus verdict, so the consensus itself needs a check against people.
We drew $400$ conversations, $25$ from each of the sixteen conditions, cycling through the eleven models inside each condition so that every model is represented.
The draw is not tilted towards hard cases.
It does not need to be: the eleven judges agree unanimously on only $27.8\%$ of conversations, so disagreement is already the common case.
Two annotators labelled every conversation independently.
Each saw the question, the reference answer, and the full conversation with its images, but never the judges' votes.
The instruction was a single decision: does the model's final answer mean the same as the reference answer?
A partial answer counts as wrong, an answer the model later withdrew does not count, and a conversation with no answer is wrong.
The annotators agreed on 97.5\% of conversations ($\kappa = 0.95$), and the remaining cases were settled by discussion.
\textit{\textbf{The consensus agrees with people.}}
Against the adjudicated human verdict, the consensus is correct on 96.0\% of conversations ($\kappa = 0.92$).
The best single judge reaches 92.5\%.
\textit{\textbf{Voting is what earns the gap.}}
A single judge carries its own leniency and its own preference for itself, Appendix~\ref{app:judges}.
A majority of eleven cancels most of both.
Agreement is 99.1\% on conversations where the judges were unanimous and 94.8\% where they split, so the remaining errors sit where the judges themselves were unsure.

%% file: tables/category.tex
\begin{wraptable}{r}{0.5\textwidth}
\centering
\vspace{-4pt}

\caption{\textbf{Accuracy by question type and delivery condition.}}
\label{tab:category}

\footnotesize
\setlength{\tabcolsep}{1.5pt}
\renewcommand{\arraystretch}{1.02}

\resizebox{\linewidth}{!}{%
\begin{tabular}{@{}l r *{8}{c}@{}}

\toprule

Question type
& $n$
& \rot{\textsc{full}}
& \rot{\textsc{concat-q}}
& \rot{\textsc{sharded-q}}
& \rot{\textsc{concat-v}}
& \rot{\textsc{sharded-v}}
& \rot{\textsc{recap-v}}
& \rot{\textsc{all-lyr}}
& \rot{\textsc{oracle}}
\\

\midrule

Direct ask
& 83
& \cellcolor{lvqaBase!30}\textbf{.663}
& \cellcolor{lvqaDown!8}.632
& \cellcolor{lvqaDown!36}.513
& \cellcolor{lvqaDown!30}.538
& \cellcolor{lvqaDown!60}.376
& \cellcolor{lvqaUp!6}.693
& \cellcolor{lvqaDown!3}.650
& \cellcolor{lvqaDown!22}.573
\\

Attribute binding
& 65
& \cellcolor{lvqaBase!30}\textbf{.601}
& \cellcolor{lvqaDown!9}.565
& \cellcolor{lvqaDown!40}.434
& \cellcolor{lvqaDown!23}.507
& \cellcolor{lvqaDown!49}.398
& \cellcolor{lvqaUp!17}.682
& \cellcolor{lvqaUp!1}.604
& \cellcolor{lvqaDown!12}.549
\\

Occlusion
& 17
& \cellcolor{lvqaBase!30}\textbf{.594}
& \cellcolor{lvqaDown!1}.590
& \cellcolor{lvqaDown!44}.412
& \cellcolor{lvqaDown!59}.346
& \cellcolor{lvqaDown!60}.248
& \cellcolor{lvqaUp!6}.624
& \cellcolor{lvqaDown!8}.560
& \cellcolor{lvqaDown!36}.443
\\

Counting
& 44
& \cellcolor{lvqaBase!30}\textbf{.642}
& \cellcolor{lvqaDown!25}.539
& \cellcolor{lvqaDown!36}.492
& \cellcolor{lvqaDown!60}.298
& \cellcolor{lvqaDown!60}.235
& \cellcolor{lvqaDown!0}.641
& \cellcolor{lvqaDown!9}.604
& \cellcolor{lvqaDown!60}.342
\\

Spatial
& 54
& \cellcolor{lvqaBase!30}\textbf{.524}
& \cellcolor{lvqaDown!3}.513
& \cellcolor{lvqaUp!1}.527
& \cellcolor{lvqaDown!59}.280
& \cellcolor{lvqaDown!60}.252
& \cellcolor{lvqaUp!28}.658
& \cellcolor{lvqaDown!7}.496
& \cellcolor{lvqaDown!53}.305
\\

Multi-hop
& 37
& \cellcolor{lvqaBase!30}\textbf{.550}
& \cellcolor{lvqaDown!13}.495
& \cellcolor{lvqaDown!60}.293
& \cellcolor{lvqaDown!43}.372
& \cellcolor{lvqaDown!60}.282
& \cellcolor{lvqaUp!13}.615
& \cellcolor{lvqaDown!1}.546
& \cellcolor{lvqaDown!37}.394
\\

\midrule

\textit{Single-object}
& 165
& \cellcolor{lvqaBase!30}\textbf{.632}
& \cellcolor{lvqaDown!7}-.030
& \cellcolor{lvqaDown!38}-.160
& \cellcolor{lvqaDown!30}-.126
& \cellcolor{lvqaDown!60}-.260
& \cellcolor{lvqaUp!10}+.050
& \cellcolor{lvqaDown!2}-.009
& \cellcolor{lvqaDown!20}-.081
\\

\textit{Relational}
& 135
& \cellcolor{lvqaBase!30}\textbf{.570}
& \cellcolor{lvqaDown!13}-.053
& \cellcolor{lvqaDown!28}-.118
& \cellcolor{lvqaDown!60}-.259
& \cellcolor{lvqaDown!60}-.315
& \cellcolor{lvqaUp!15}+.071
& \cellcolor{lvqaDown!6}-.025
& \cellcolor{lvqaDown!55}-.228
\\

\bottomrule
\end{tabular}%
}

\vspace{-5mm}
\end{wraptable}

%% file: tables/first-attempt.tex
\begin{wraptable}{r}{0.5\textwidth}
\centering
\vspace{-3mm}
\caption{\textbf{When models first attempt an answer.}}
\label{tab:first-answer-attempt}
\small
\setlength{\tabcolsep}{3pt}
\renewcommand{\arraystretch}{1.05}

\resizebox{\linewidth}{!}{%
\begin{tabular}{l ccccc c}
\toprule
& \multicolumn{5}{c}{\textsc{first answer attempt}}
& \multicolumn{1}{c}{\textsc{commits}} \\
\cmidrule(lr){2-6}
\cmidrule(lr){7-7}

Model
& earliest
& early
& midway
& late
& latest
& share \\

&
{\tiny 0--20\%}
& {\tiny 20--40\%}
& {\tiny 40--60\%}
& {\tiny 60--80\%}
& {\tiny 80--100\%}
& \\

\midrule

Q2.5-VL-3B
& \cellcolor{lvqaDown!30}\nb{.206}
& \cellcolor{lvqaDown!7}\nb{.220}
& \cellcolor{lvqaUp!11}\nb{.232}
& \cellcolor{lvqaUp!52}\nb{.261}
& \cellcolor{lvqaDown!35}\nb{.203}
& 51\% \\

Q3-VL-4B
& \cellcolor{lvqaDown!60}\nb{.417}
& \cellcolor{lvqaUp!31}\nb{.561}
& \cellcolor{lvqaUp!22}\nb{.545}
& \cellcolor{lvqaUp!4}\nb{.514}
& \cellcolor{lvqaDown!5}\nb{.500}
& 63\% \\

Q2.5-VL-7B
& \cellcolor{lvqaDown!60}\nb{.332}
& \cellcolor{lvqaUp!33}\nb{.393}
& \cellcolor{lvqaUp!40}\nb{.398}
& \cellcolor{lvqaUp!29}\nb{.390}
& \cellcolor{lvqaDown!58}\nb{.333}
& 53\% \\

Q3-VL-8B
& \cellcolor{lvqaDown!60}\nb{.502}
& \cellcolor{lvqaUp!13}\nb{.626}
& \cellcolor{lvqaUp!29}\nb{.657}
& \cellcolor{lvqaUp!24}\nb{.646}
& \cellcolor{lvqaDown!16}\nb{.574}
& 45\% \\

IVL3.5-8B
& \cellcolor{lvqaDown!60}\nb{.357}
& \cellcolor{lvqaUp!30}\nb{.442}
& \cellcolor{lvqaUp!28}\nb{.440}
& \cellcolor{lvqaDown!6}\nb{.405}
& \cellcolor{lvqaDown!1}\nb{.410}
& 45\% \\

Min-3-8B
& \cellcolor{lvqaDown!55}\nb{.200}
& \cellcolor{lvqaUp!2}\nb{.250}
& \cellcolor{lvqaUp!18}\nb{.267}
& \cellcolor{lvqaUp!52}\nb{.302}
& \cellcolor{lvqaDown!28}\nb{.224}
& 47\% \\

Q3.5-9B
& \cellcolor{lvqaDown!60}\nb{.439}
& \cellcolor{lvqaUp!0}\nb{.530}
& \cellcolor{lvqaDown!0}\nb{.530}
& \cellcolor{lvqaUp!10}\nb{.548}
& \cellcolor{lvqaUp!42}\nb{.603}
& 58\% \\

Pixtral-12B
& \cellcolor{lvqaDown!25}\nb{.341}
& \cellcolor{lvqaDown!33}\nb{.337}
& \cellcolor{lvqaUp!52}\nb{.388}
& \cellcolor{lvqaUp!45}\nb{.384}
& \cellcolor{lvqaDown!54}\nb{.326}
& 41\% \\

Q3.6-27B
& \cellcolor{lvqaDown!60}\nb{.498}
& \cellcolor{lvqaUp!52}\nb{.577}
& \cellcolor{lvqaUp!16}\nb{.550}
& \cellcolor{lvqaUp!23}\nb{.556}
& \cellcolor{lvqaDown!46}\nb{.508}
& 44\% \\

Q3-VL-30B
& \cellcolor{lvqaDown!60}\nb{.521}
& \cellcolor{lvqaUp!15}\nb{.598}
& \cellcolor{lvqaUp!40}\nb{.626}
& \cellcolor{lvqaUp!24}\nb{.608}
& \cellcolor{lvqaDown!32}\nb{.549}
& 54\% \\

Q2.5-VL-32B
& \cellcolor{lvqaDown!60}\nb{.188}
& \cellcolor{lvqaUp!10}\nb{.457}
& \cellcolor{lvqaUp!21}\nb{.504}
& \cellcolor{lvqaUp!28}\nb{.533}
& \cellcolor{lvqaDown!9}\nb{.378}
& 42\% \\

GPT-5.6-L$^{\dagger}$
& \cellcolor{lvqaDown!60}\nb{.587}
& \cellcolor{lvqaDown!15}\nb{.622}
& \cellcolor{lvqaUp!26}\nb{.655}
& \cellcolor{lvqaUp!14}\nb{.645}
& \cellcolor{lvqaUp!25}\nb{.655}
& 36\% \\

Gemini-3.7-F$^{\dagger}$
& \cellcolor{lvqaUp!52}\nb{.865}
& \cellcolor{lvqaDown!38}\nb{.829}
& \cellcolor{lvqaDown!43}\nb{.828}
& \cellcolor{lvqaDown!18}\nb{.837}
& \cellcolor{lvqaUp!33}\nb{.857}
& \cellcolor{white}34\% \\

\midrule

\textit{Mean, Open}
& \nb{.388}
& \nb{.458}
& \nb{.460}
& \nb{.473}
& \nb{.403}
& \\

\bottomrule
\end{tabular}%
}
\vspace{-5mm}
\end{wraptable}

%% file: tables/verbosity.tex
\begin{table*}[h]
\centering

\caption{\textbf{Verbosity: Accuracy by evidence length and delivery mode.}
Models are grouped by the length of the evidence they receive, either in one turn or spread across multiple turns.}
\label{tab:verbosity}

\scriptsize
\setlength{\tabcolsep}{1.2pt}
\renewcommand{\arraystretch}{1.02}

% \resizebox{\textwidth}{!}{%
\begin{tabular}{@{}l ccccc ccccc@{}}

\toprule
&
\multicolumn{5}{c}{\textsc{evidence in one turn}}
&
\multicolumn{5}{c}{\textsc{evidence across turns}}
\\

\cmidrule(lr){2-6}
\cmidrule(lr){7-11}

Model
& \rot{\textsc{shortest}}
& \rot{\textsc{short}}
& \rot{\textsc{median}}
& \rot{\textsc{long}}
& \rot{\textsc{longest}}
& \rot{\textsc{shortest}}
& \rot{\textsc{short}}
& \rot{\textsc{median}}
& \rot{\textsc{long}}
& \rot{\textsc{longest}}
\\

\midrule

Q2.5-VL-3B
& \cellcolor{lvqaDown!14}\nb{.311}
& \cellcolor{lvqaUp!24}\nb{.365}
& \cellcolor{lvqaUp!13}\nb{.349}
& \cellcolor{lvqaUp!27}\nb{.369}
& \cellcolor{lvqaDown!60}\nb{.253}
& \cellcolor{lvqaUp!52}\nb{.270}
& \cellcolor{lvqaUp!16}\nb{.249}
& \cellcolor{lvqaDown!26}\nb{.227}
& \cellcolor{lvqaDown!8}\nb{.236}
& \cellcolor{lvqaDown!45}\nb{.218}
\\

Q3-VL-4B
& \cellcolor{lvqaUp!4}\nb{.649}
& \cellcolor{lvqaUp!21}\nb{.706}
& \cellcolor{lvqaUp!23}\nb{.715}
& \cellcolor{lvqaUp!4}\nb{.647}
& \cellcolor{lvqaDown!60}\nb{.451}
& \cellcolor{lvqaUp!32}\nb{.592}
& \cellcolor{lvqaUp!16}\nb{.569}
& \cellcolor{lvqaUp!12}\nb{.564}
& \cellcolor{lvqaDown!10}\nb{.534}
& \cellcolor{lvqaDown!60}\nb{.474}
\\

Q2.5-VL-7B
& \cellcolor{lvqaUp!4}\nb{.478}
& \cellcolor{lvqaUp!39}\nb{.565}
& \cellcolor{lvqaUp!34}\nb{.554}
& \cellcolor{lvqaDown!29}\nb{.403}
& \cellcolor{lvqaDown!60}\nb{.336}
& \cellcolor{lvqaUp!19}\nb{.412}
& \cellcolor{lvqaUp!19}\nb{.412}
& \cellcolor{lvqaUp!21}\nb{.415}
& \cellcolor{lvqaDown!9}\nb{.381}
& \cellcolor{lvqaDown!60}\nb{.327}
\\

Q3-VL-8B
& \cellcolor{lvqaUp!12}\nb{.696}
& \cellcolor{lvqaUp!5}\nb{.676}
& \cellcolor{lvqaUp!25}\nb{.732}
& \cellcolor{lvqaUp!11}\nb{.694}
& \cellcolor{lvqaDown!60}\nb{.521}
& \cellcolor{lvqaUp!22}\nb{.649}
& \cellcolor{lvqaUp!28}\nb{.657}
& \cellcolor{lvqaUp!1}\nb{.623}
& \cellcolor{lvqaUp!0}\nb{.622}
& \cellcolor{lvqaDown!60}\nb{.557}
\\

IVL3.5-8B
& \cellcolor{lvqaUp!20}\nb{.524}
& \cellcolor{lvqaDown!17}\nb{.468}
& \cellcolor{lvqaUp!13}\nb{.512}
& \cellcolor{lvqaUp!34}\nb{.546}
& \cellcolor{lvqaDown!60}\nb{.407}
& \cellcolor{lvqaUp!33}\nb{.458}
& \cellcolor{lvqaUp!2}\nb{.431}
& \cellcolor{lvqaDown!60}\nb{.385}
& \cellcolor{lvqaDown!5}\nb{.426}
& \cellcolor{lvqaUp!21}\nb{.448}
\\

Min-3-8B
& \cellcolor{lvqaUp!39}\nb{.526}
& \cellcolor{lvqaUp!37}\nb{.519}
& \cellcolor{lvqaUp!14}\nb{.446}
& \cellcolor{lvqaDown!43}\nb{.276}
& \cellcolor{lvqaDown!60}\nb{.229}
& \cellcolor{lvqaUp!29}\nb{.297}
& \cellcolor{lvqaUp!22}\nb{.291}
& \cellcolor{lvqaUp!32}\nb{.300}
& \cellcolor{lvqaDown!37}\nb{.242}
& \cellcolor{lvqaDown!60}\nb{.224}
\\

Q3.5-9B
& \cellcolor{lvqaUp!19}\nb{.617}
& \cellcolor{lvqaDown!2}\nb{.576}
& \cellcolor{lvqaUp!10}\nb{.600}
& \cellcolor{lvqaUp!26}\nb{.631}
& \cellcolor{lvqaDown!60}\nb{.479}
& \cellcolor{lvqaUp!32}\nb{.563}
& \cellcolor{lvqaUp!31}\nb{.562}
& \cellcolor{lvqaUp!5}\nb{.531}
& \cellcolor{lvqaDown!18}\nb{.506}
& \cellcolor{lvqaDown!60}\nb{.462}
\\

Pixtral-12B
& \cellcolor{lvqaDown!26}\nb{.453}
& \cellcolor{lvqaUp!27}\nb{.547}
& \cellcolor{lvqaUp!7}\nb{.510}
& \cellcolor{lvqaUp!41}\nb{.575}
& \cellcolor{lvqaDown!60}\nb{.396}
& \cellcolor{lvqaUp!38}\nb{.401}
& \cellcolor{lvqaUp!11}\nb{.383}
& \cellcolor{lvqaUp!4}\nb{.379}
& \cellcolor{lvqaDown!2}\nb{.375}
& \cellcolor{lvqaDown!60}\nb{.341}
\\

Q3.6-27B
& \cellcolor{lvqaDown!1}\nb{.639}
& \cellcolor{lvqaDown!8}\nb{.625}
& \cellcolor{lvqaUp!29}\nb{.704}
& \cellcolor{lvqaUp!31}\nb{.710}
& \cellcolor{lvqaDown!60}\nb{.526}
& \cellcolor{lvqaUp!33}\nb{.609}
& \cellcolor{lvqaUp!24}\nb{.598}
& \cellcolor{lvqaUp!9}\nb{.578}
& \cellcolor{lvqaDown!17}\nb{.548}
& \cellcolor{lvqaDown!60}\nb{.499}
\\

Q3-VL-30B
& \cellcolor{lvqaUp!3}\nb{.667}
& \cellcolor{lvqaUp!8}\nb{.678}
& \cellcolor{lvqaUp!45}\nb{.764}
& \cellcolor{lvqaDown!4}\nb{.651}
& \cellcolor{lvqaDown!60}\nb{.539}
& \cellcolor{lvqaDown!20}\nb{.584}
& \cellcolor{lvqaUp!52}\nb{.644}
& \cellcolor{lvqaUp!29}\nb{.624}
& \cellcolor{lvqaDown!30}\nb{.577}
& \cellcolor{lvqaDown!43}\nb{.566}
\\

Q2.5-VL-32B
& \cellcolor{lvqaUp!16}\nb{.579}
& \cellcolor{lvqaUp!28}\nb{.618}
& \cellcolor{lvqaUp!20}\nb{.590}
& \cellcolor{lvqaDown!13}\nb{.485}
& \cellcolor{lvqaDown!60}\nb{.347}
& \cellcolor{lvqaUp!51}\nb{.589}
& \cellcolor{lvqaUp!18}\nb{.526}
& \cellcolor{lvqaDown!3}\nb{.488}
& \cellcolor{lvqaDown!16}\nb{.467}
& \cellcolor{lvqaDown!60}\nb{.396}
\\

GPT-5.6-L$^{\dagger}$
& \cellcolor{lvqaUp!52}\nb{.758}
& \cellcolor{lvqaUp!39}\nb{.739}
& \cellcolor{lvqaDown!38}\nb{.629}
& \cellcolor{lvqaDown!23}\nb{.649}
& \cellcolor{lvqaDown!44}\nb{.622}
& \cellcolor{lvqaUp!39}\nb{.697}
& \cellcolor{lvqaUp!29}\nb{.684}
& \cellcolor{lvqaDown!1}\nb{.644}
& \cellcolor{lvqaDown!18}\nb{.624}
& \cellcolor{lvqaDown!60}\nb{.575}
\\

Gemini-3.7-F$^{\dagger}$
& \cellcolor{lvqaDown!60}\nb{.815}
& \cellcolor{lvqaUp!20}\nb{.875}
& \cellcolor{lvqaUp!18}\nb{.873}
& \cellcolor{lvqaUp!48}\nb{.898}
& \cellcolor{lvqaDown!40}\nb{.829}
& \cellcolor{lvqaUp!41}\nb{.858}
& \cellcolor{lvqaUp!52}\nb{.864}
& \cellcolor{lvqaDown!24}\nb{.828}
& \cellcolor{lvqaDown!29}\nb{.826}
& \cellcolor{lvqaDown!53}\nb{.816}
\\

\midrule

\textit{Mean, Open}
& \nb{.558}
& \nb{.577}
& \nb{.589}
& \nb{.544}
& \nb{.408}
& \nb{.493}
& \nb{.484}
& \nb{.465}
& \nb{.447}
& \nb{.410}
\\

\bottomrule

\end{tabular}%
% }

\end{table*}

%% file: tables/ungrounded-correctness.tex
\begin{wraptable}{r}{0.5\textwidth}
\centering
\vspace{-3mm}
% \caption{\textbf{Ungrounded responses and grounded accuracy.}}
\caption{\textbf{Ungrounded answers remain common across delivery settings.}
\textbf{\textit{Left}}: reports the share of responses produced before the required evidence was available; \textbf{\textit{Right}}: compares standard accuracy with accuracy restricted to grounded answers.}
\label{tab:ungrounded-grounded}

\scriptsize
\setlength{\tabcolsep}{2.8pt}
\renewcommand{\arraystretch}{1.02}

\resizebox{\linewidth}{!}{%
\begin{tabular}{l ccc cc}

\toprule

&
\multicolumn{3}{c}{\textsc{ungrounded share}}
&
\multicolumn{2}{c}{\textsc{accuracy}}
\\

\cmidrule(lr){2-4}
\cmidrule(lr){5-6}

Model
& Visual
& Layered
& Order
& as scored
& grounded only
\\

\midrule

Q2.5-VL-3B
& \cellcolor{lvqaDown!51}\nb{.603}
& \cellcolor{lvqaDown!41}\nb{.556}
& \cellcolor{lvqaDown!52}\nb{.606}
& \nb{.223}
& \textbf{\nb{.090}}
\\

Q3-VL-4B
& \cellcolor{lvqaDown!6}\nb{.397}
& \cellcolor{lvqaDown!0}\nb{.369}
& \cellcolor{lvqaUp!18}\nb{.274}
& \nb{.482}
& \textbf{\nb{.312}}
\\

Q2.5-VL-7B
& \cellcolor{lvqaDown!19}\nb{.454}
& \cellcolor{lvqaDown!8}\nb{.404}
& \cellcolor{lvqaDown!2}\nb{.376}
& \nb{.338}
& \textbf{\nb{.196}}
\\

Q3-VL-8B
& \cellcolor{lvqaUp!8}\nb{.326}
& \cellcolor{lvqaUp!15}\nb{.291}
& \cellcolor{lvqaUp!28}\nb{.221}
& \nb{.573}
& \textbf{\nb{.411}}
\\

IVL3.5-8B
& \cellcolor{lvqaDown!24}\nb{.478}
& \cellcolor{lvqaUp!20}\nb{.260}
& \cellcolor{lvqaDown!2}\nb{.377}
& \nb{.403}
& \textbf{\nb{.243}}
\\

Min-3-8B
& \cellcolor{lvqaUp!5}\nb{.344}
& \cellcolor{lvqaDown!16}\nb{.443}
& \cellcolor{lvqaUp!19}\nb{.269}
& \nb{.270}
& \textbf{\nb{.181}}
\\

Q3.5-9B
& \cellcolor{lvqaUp!1}\nb{.362}
& \cellcolor{lvqaDown!10}\nb{.413}
& \cellcolor{lvqaUp!21}\nb{.257}
& \nb{.473}
& \textbf{\nb{.315}}
\\

Pixtral-12B
& \cellcolor{lvqaDown!16}\nb{.443}
& \cellcolor{lvqaDown!17}\nb{.445}
& \cellcolor{lvqaDown!6}\nb{.398}
& \nb{.339}
& \textbf{\nb{.193}}
\\

Q3.6-27B
& \cellcolor{lvqaDown!7}\nb{.399}
& \cellcolor{lvqaDown!3}\nb{.382}
& \cellcolor{lvqaUp!12}\nb{.306}
& \nb{.524}
& \textbf{\nb{.333}}
\\

Q3-VL-30B
& \cellcolor{lvqaDown!16}\nb{.443}
& \cellcolor{lvqaUp!8}\nb{.326}
& \cellcolor{lvqaUp!7}\nb{.330}
& \nb{.556}
& \textbf{\nb{.344}}
\\

Q2.5-VL-32B
& \cellcolor{lvqaUp!33}\nb{.192}
& \cellcolor{lvqaUp!8}\nb{.325}
& \cellcolor{lvqaUp!52}\nb{.093}
& \nb{.434}
& \textbf{\nb{.354}}
\\

GPT-5.6-L$^{\dagger}$
& \cellcolor{lvqaUp!4}\nb{.347}
& \cellcolor{lvqaDown!1}\nb{.371}
& \cellcolor{lvqaUp!32}\nb{.198}
& \nb{.630}
& \textbf{\nb{.440}}
\\

Gemini-3.7-F$^{\dagger}$
& \cellcolor{lvqaUp!18}\nb{.273}
& \cellcolor{lvqaDown!0}\nb{.369}
& \cellcolor{lvqaUp!48}\nb{.116}
& \nb{.835}
& \textbf{\nb{.635}}
\\

\midrule

\textit{Mean, Open}
& \nb{.393}
& \nb{.362}
& \nb{.298}
& \nb{.420}
& \textbf{\nb{.270}}
\\

\bottomrule
\end{tabular}%
}
\vspace{-5mm}
\end{wraptable}

%% file: sections/appendix-additional-1.tex
\section{Ruling out the alternatives}
\label{app:alternatives}

The scene loss has three obvious rivals.
It could be missing context, because a layer shows less of the world than the composite.
It could be distribution shift, because a lone object on a blank canvas is a strange input.
It could be an artifact of how finely we cut a scene.
Each subsection takes one and tests it against the data.
The last asks whether the relational gap survives uncertainty measured at the scene level.

\subsection{Is the oracle loss about missing context, or missing composition?}
\label{app:oracle-why}

\textsc{oracle} removes two things at once: the surrounding scene, and the composed view.
\textsc{all-layers} keeps the composite and adds every layer beside it.
It costs $-.017$.
\textsc{oracle} takes the composite away and keeps only the gold layers.
It costs $-.148$.
Adding layers is nearly free when the composite is there, so the loss tracks the composite, not the layer count.
The second comparison uses the annotation.
For $97$ questions the background is part of the gold set, so \textsc{oracle} delivers scene context.
For the other $203$ it does not.
Context does not rescue the model.
Questions whose oracle includes the background lose $-.171$, and questions whose oracle excludes it lose $-.136$.
These are different questions, so this is a split rather than a control, but it points the wrong way for the context explanation.
What \textsc{oracle} destroys is the arrangement.
The layers preserve every object at its original position, and the model still cannot read the relations between them.

\subsection{Is the loss just distribution shift from layer images?}
\label{app:shift}

A layer is an unusual input.
It is one object on a blank canvas, and models rarely see that in training.
If the loss were distribution shift, then feeding layers at all would hurt.
\textsc{all-layers} feeds exactly the same layer renders, in the same number, with the composite alongside.
It costs $-.017$ against $-.187$ for the same layers without the composite.
The renders are therefore not the problem.
Whether the composed view is present is.

\subsection{Does the result depend on how finely a scene is decomposed?}
\label{app:depth}

Our scenes decompose into $2$ to $6$ layers, so the benchmark contains its own sensitivity analysis.
\textsc{full} is flat at about $.60$ across depth, so deeper scenes are not harder scenes.
Both losses grow steadily with depth, from $-.050$ to $-.269$.
The effect is not an artifact of one decomposition setting.
It is a dose response: the more pieces a scene is cut into, the more of it is lost.
\input{tables/depth}

\subsection{Are the relational losses significant at the scene level?}
\label{app:relational-ci}
Both headline losses are about three times larger on questions about relations between objects than on questions about one object.
We test the difference directly, with a scene-level cluster bootstrap over $2000$ draws.
\input{tables/relational-loss}
Each draw resamples the $93$ scenes, keeps all of a scene's questions, and recomputes both losses and their difference inside the replicate.
Both intervals exclude zero.
Fragmenting a scene and cropping to an oracle are different operations, and they cost the same kind of question the same extra amount.
That is the mechanism: both keep the objects and destroy what holds between them.

\subsection{Where the loss falls, and what putting it back buys}
\label{app:shape}

Two questions the headline averages cannot answer.
Is the loss spread across the benchmark or concentrated in part of it?
And when recomposition recovers accuracy, what exactly is it recovering?

\begin{figure}[h]
\centering
\includegraphics[width=0.75\linewidth]{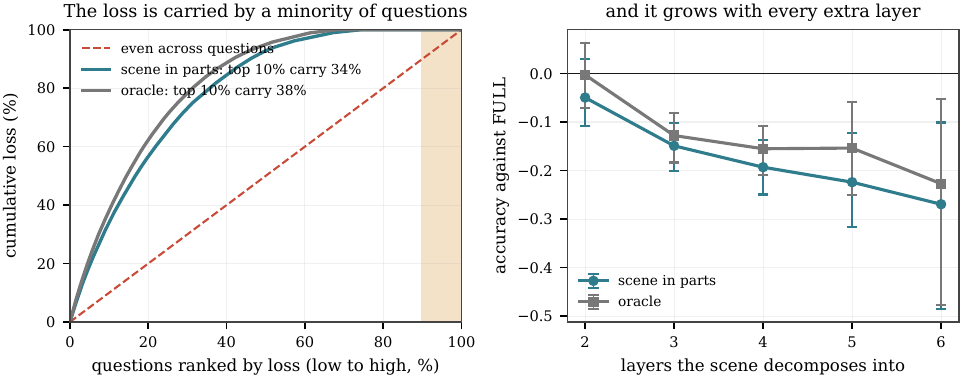}
\caption{\textbf{The loss is concentrated, and it scales with depth.}
\textit{Left:} questions ranked by loss against the running total of that loss.
\textit{Right:} both losses by the number of layers a scene decomposes into, with $95\%$ scene-level bootstrap intervals.}
\label{fig:concentration}
\end{figure}

\textbf{\textit{A minority of questions carries the loss.}}
Figure~\ref{fig:concentration} ranks questions by how much they lose and plots the running total.
An even loss would trace the diagonal.
The measured curve does not come close.
The worst tenth of questions carries $34\%$ of the total loss from fragmenting the scene, the worst fifth carries $57\%$, and $78$ of the $300$ questions lose nothing at all.
The oracle is more concentrated still, at $38\%$.
Fragmentation does not make every question slightly harder.
It destroys some questions and leaves the rest alone, which is what a compositional bottleneck looks like and not what uniform information loss looks like.
\textbf{\textit{Each extra layer costs more.}}
The right panel splits the same two losses by how finely the scene decomposes.
\textsc{full} is flat across depth at about $.60$, so deeper scenes are not harder scenes.
Both losses grow monotonically, from $-.050$ at two layers to $-.269$ at six.
The cost is not a fixed penalty for using layers at all.
It scales with how much of the scene was taken apart.

\begin{figure}[h]
\centering
\includegraphics[width=0.75\linewidth]{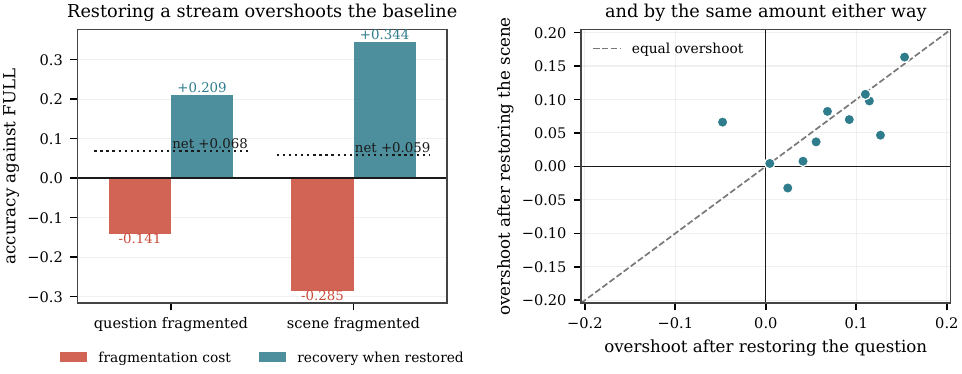}
\caption{\textbf{Restoring a fragmented stream overshoots the baseline.}
\textit{Left:} the cost of fragmenting each stream and the recovery when it is restored, against \textsc{full}.
\textit{Right:} the two overshoots per model. Points on the diagonal mean the final turn is worth the same either way.}
\label{fig:recomposition}
\end{figure}

\textbf{\textit{Recomposition pays twice.}}
Figure~\ref{fig:recomposition} decomposes what a final restoring turn is worth.
Fragmenting the question costs $-.141$ and restoring it recovers $+.209$.
Fragmenting the scene costs twice as much at $-.285$, and restoring it recovers $+.344$.
Both end above the \textsc{full} baseline, and by almost the same margin: $+.068$ and $+.059$.
The right panel shows the two overshoots against each other, one point per model, sitting on the diagonal.
So the recovery has two parts.
A final turn that restates everything is worth about $+.06$ whatever was fragmented, which is the value of a clean summary at the end of a conversation.
The rest is undoing the fragmentation, and that part is twice as large for the scene as for the question.
This also settles a smaller puzzle: \textsc{recap-v} scoring above \textsc{full} is not a quirk of the visual stream, since \textsc{recap-q} does the same thing by the same amount.

%% file: tables/depth.tex
% \begin{wraptable}{r}{0.5\textwidth}
% \centering
% \vspace{-5mm}
% \caption{\textbf{Loss by scene depth.}
% % Both losses grow with the number of layers, and neither depends on a particular granularity.
% }
% \label{tab:depth}
% \footnotesize
% \setlength{\tabcolsep}{2pt}
% \renewcommand{\arraystretch}{1.05}
% \begin{tabular}{@{}ccccc@{}}
% \toprule
% Layers
% & Questions
% & \textsc{full}
% & $\Delta$ \textsc{concat-v}
% & $\Delta$ \textsc{oracle}
% \\
% \midrule
% 2 &  11 & .606 & $-.050$ & $-.003$ \\
% 3 & 100 & .597 & $-.149$ & $-.128$ \\
% 4 & 110 & .611 & $-.193$ & $-.155$ \\
% 5 &  51 & .594 & $-.224$ & $-.156$ \\
% 6 &  28 & .616 & $-.269$ & $-.229$ \\
% \bottomrule
% \end{tabular}
% \vspace{-7mm}
% \end{wraptable}

\begin{table}[h]
\centering
\caption{\textbf{Loss by scene depth.}
Both losses grow with the number of layers, and neither depends on a particular granularity.}
\label{tab:depth}
\footnotesize
\setlength{\tabcolsep}{4pt}
\renewcommand{\arraystretch}{1.05}
\begin{tabular}{@{}ccccc@{}}
\toprule
Layers & Questions & \textsc{full} & $\Delta$ \textsc{concat-v} & $\Delta$ \textsc{oracle} \\
\midrule
2 &  11 & .606 & $-.050$ & $-.003$ \\
3 & 100 & .597 & $-.149$ & $-.128$ \\
4 & 110 & .611 & $-.193$ & $-.155$ \\
5 &  51 & .594 & $-.224$ & $-.156$ \\
6 &  28 & .616 & $-.269$ & $-.229$ \\
\bottomrule
\end{tabular}
\end{table}

%% file: tables/relational-loss.tex
\begin{wraptable}{r}{0.5\textwidth}
\centering
\vspace{-6pt}

\caption{\textbf{Relational questions lose more under visual decomposition.}
Extra loss is the difference between relational and single-object degradation.}
\label{tab:relational-loss}

\footnotesize
\setlength{\tabcolsep}{3pt}
\renewcommand{\arraystretch}{1.05}

\resizebox{\linewidth}{!}{%
\begin{tabular}{@{}lcccc@{}}
\toprule
Condition & Relational & Single-object & Extra loss & $95\%$ CI \\
\midrule
\textsc{concat-v} & $-.259$ & $-.126$ & $-.133$ & $[-.189, -.074]$ \\
\textsc{oracle}   & $-.228$ & $-.081$ & $-.147$ & $[-.202, -.089]$ \\
\bottomrule
\end{tabular}%
}

\vspace{-6pt}
\end{wraptable}

%% file: sections/appendix-example.tex
\section{One question, end to end}
\label{app:example}

\begin{figure}[h]
\centering
\includegraphics[height=2.0cm]{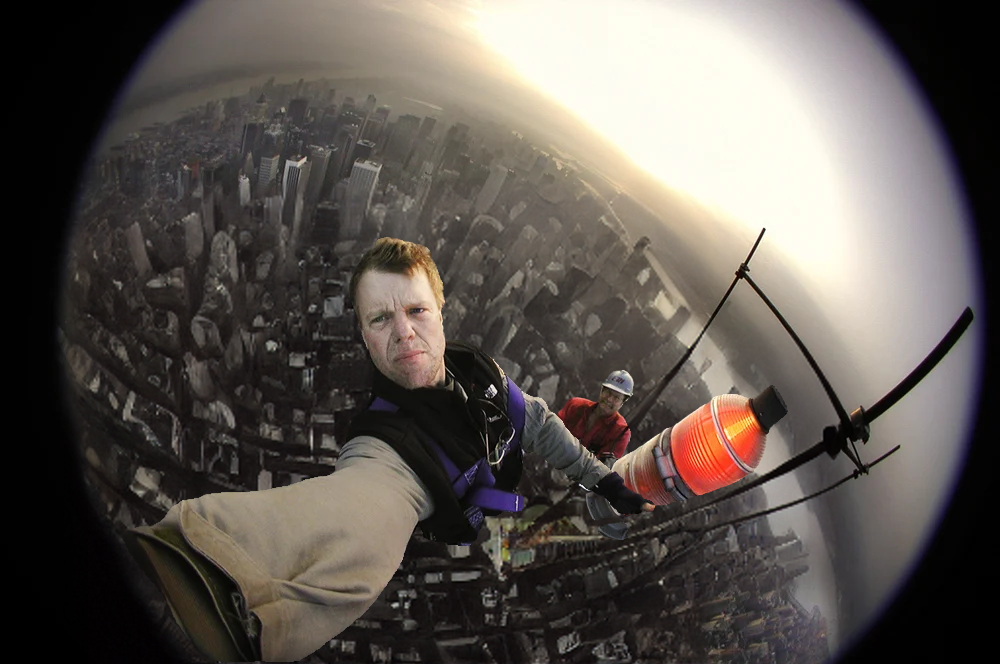}\hfill
\includegraphics[height=2.0cm]{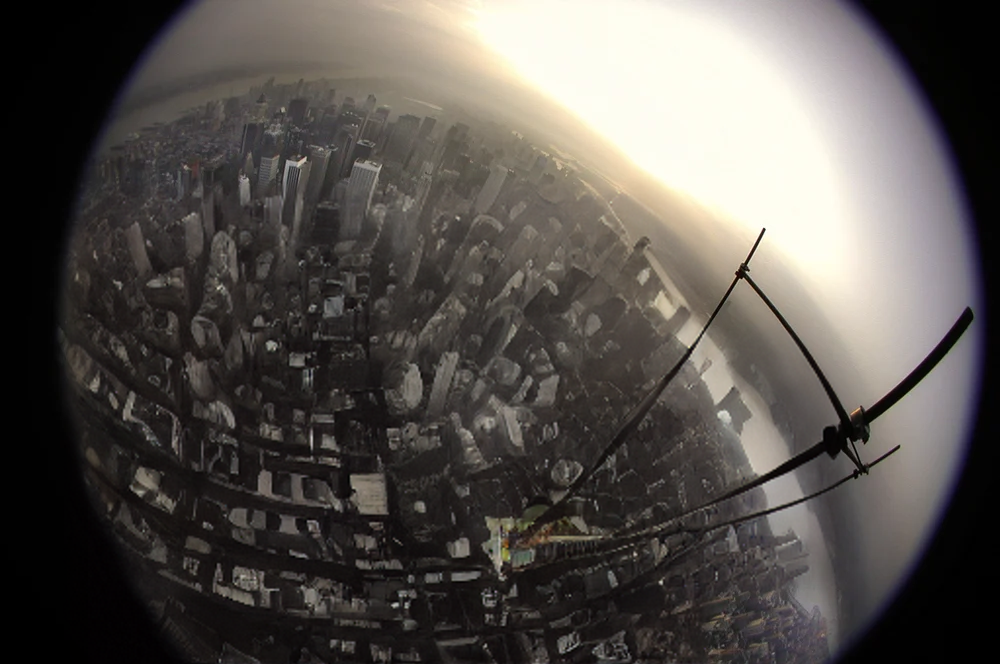}\hfill
\includegraphics[height=2.0cm]{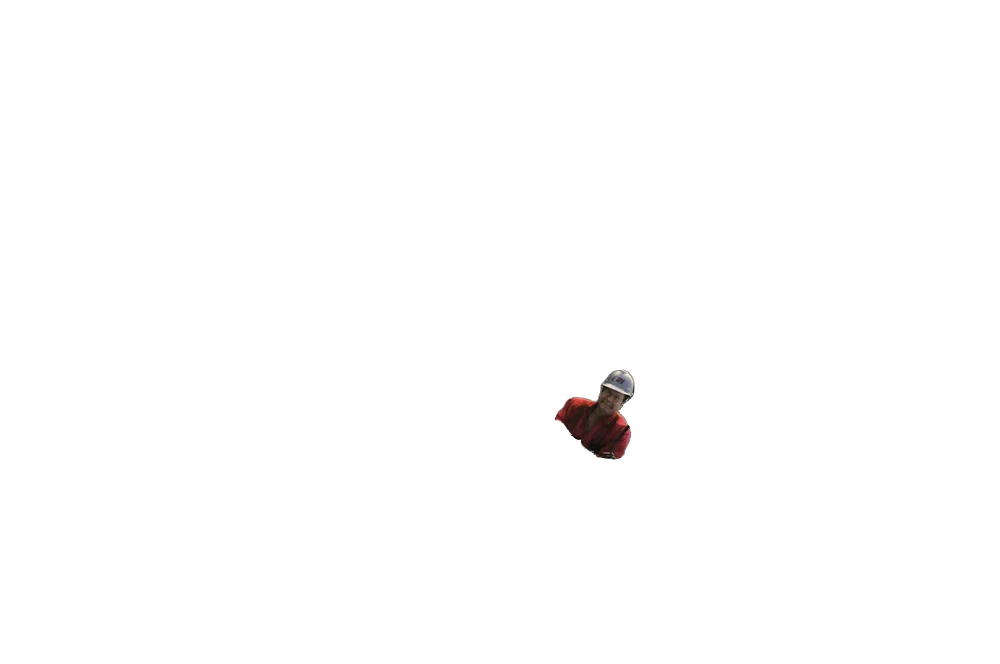}\hfill
\includegraphics[height=2.0cm]{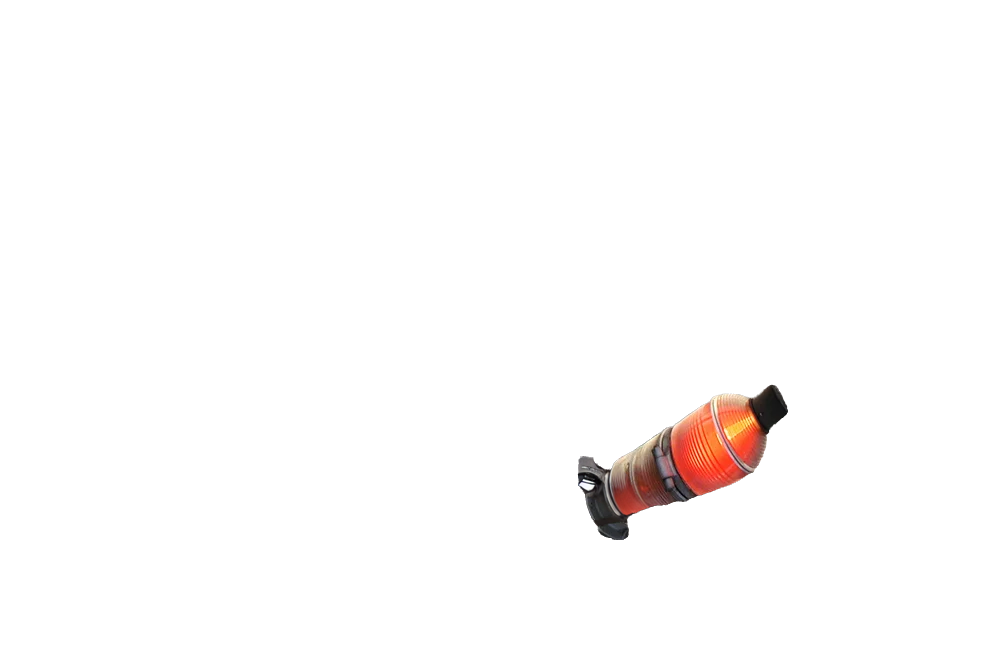}\hfill
\includegraphics[height=2.0cm]{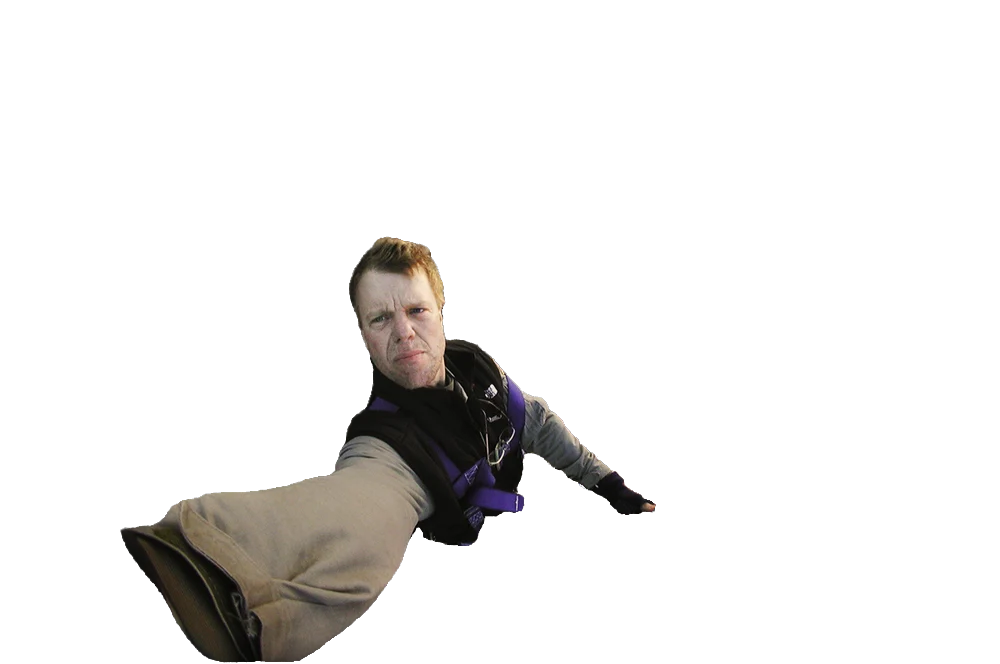}
\caption{\textbf{The scene and its four layers.}
Left to right: the composed scene, then layer $0$ (background, not needed), layer $1$ (the worker, the reference landmark), layer $2$ (the bottle, the answer), and layer $3$ (the man taking a selfie, the relational bridge).
Every layer render is used exactly as shown.}
\label{fig:example-layers}
\end{figure}

\paragraph{Q\&A.}
\emph{What object is positioned to the right of the man taking a selfie and directly in front of the worker wearing a white helmet and an orange-red shirt on top of the skyscraper?} -- \emph{a red bottle with a black cap}.

\paragraph{The question in parts.}
The sharded conditions deliver the question as six pieces, one per turn.
The first piece states the task and carries no content.
Each later piece adds one constraint, and two of them exist only to disambiguate an earlier one.
Shard $3$ is the one that makes this question hard.
It refers to \emph{the man}, and the scene contains two men.
The disambiguation arrives one turn later, in shard $4$.
A model that commits after shard $3$ is guessing which man was meant, and Section~\ref{sec:composition} shows that models commit early far more often than they should. The layers in Figure~\ref{fig:example-layers}, and shards in Table~\ref{tab:example-shards}.

\begin{table}[h]
\centering
\caption{\textbf{The six question shards, and the layers each one depends on.}
A shard with no supporting layer is asking about the question rather than the image.}
\label{tab:example-shards}
\small
\begin{tabular}{@{}clc@{}}
\toprule
Shard & Text & Supported by \\
\midrule
1 & can you help me identify a particular thing in this image? & --- \\
2 & The answer should name an object.& 2 \\
3 & The target object is positioned to the right of the man.& 2, 3 \\
4 & The referenced man is taking a selfie. & 3 \\
5 & The target object is directly in front of the worker. & 1, 2 \\
6 & The referenced worker is wearing a white helmet and an orange-red & 1 \\
  & shirt and is on top of the skyscraper.  &     \\
\bottomrule
\end{tabular}
\end{table}

\section{The same question in every condition}
\label{app:simulation}

\input{tables/example-flow}
Section~\ref{app:example} introduced one question, its scene and its four layers.
Now, that question through all sixteen conditions, Table~\ref{tab:example-schedules}.
Nothing about the question or the image changes.
Only the schedule does.
Read $Q$ as the whole question, $q_k$ as its $k$-th shard, $I$ as the composed scene, and $L_k$ as layer $k$.
The first six schedules break the question and keep the scene whole.
The next four break the scene and keep the question whole.
The next four vary the order or bind the two streams together.
The last two vary the evidence rather than the schedule.
\textsc{concat-v} against \textsc{full} holds the pixels and the turn count fixed and takes away only the composition.
\textsc{recap-v} against \textsc{sharded-v} adds nothing the model has not already seen and puts the composition back.
\textsc{oracle} against \textsc{full} gives the model exactly the three layers it needs and takes away everything else.
Note what \textsc{aligned-vq} and \textsc{misaligned-vq} do to shard $3$, the ambiguous one.
Under alignment it arrives with layer $3$, the man it refers to.
Under misalignment it arrives with layer $1$, the other man.
The pairing is the only difference between the two conditions, and the models score the same under both.

%% file: tables/example-flow.tex
\begin{table}[t]
\centering
\caption{\textbf{Sixteen delivery schedules for one question.}}
\label{tab:example-schedules}

\scriptsize
\setlength{\tabcolsep}{3pt}
\renewcommand{\arraystretch}{1.05}

\begin{tabular}{l c p{0.60\linewidth}}
\toprule
Condition & Turns & What the model receives \\
\midrule

\multicolumn{3}{l}{\textit{The question arrives in parts, the scene stays whole}} \\[2pt]

\textsc{full}
& 1
& $Q$ with $I$ \\

\textsc{concat-q}
& 1
& $q_1 q_2 q_3 q_4 q_5 q_6$ with $I$ \\

\textsc{sharded-q}
& 6
& $q_1$ with $I$; then $q_2$, $q_3$, $q_4$, $q_5$, $q_6$, one per turn \\

\textsc{recap-q}
& 7
& As \textsc{sharded-q}, then a final turn repeating $Q$ with $I$ \\

\textsc{snowball-q}
& 6
& $q_1$ with $I$; then $q_1q_2$; then $q_1q_2q_3$; and so on \\

\textsc{shuffled-concat}
& 1
& $q_1q_6q_4q_3q_2q_5$ with $I$ \\

\midrule
\multicolumn{3}{l}{\textit{The scene arrives in parts, the question stays whole}} \\[2pt]

\textsc{concat-v}
& 1
& $Q$ with $L_0$ $L_1$ $L_2$ $L_3$, no $I$ \\

\textsc{sharded-v}
& 5
& $Q$; then $L_0$, $L_1$, $L_2$, $L_3$, one per turn, no $I$ \\

\textsc{recap-v}
& 6
& As \textsc{sharded-v}, then a final turn carrying $I$ \\

\textsc{snowball-v}
& 5
& $Q$ with $L_0$; then $L_0L_1$; then $L_0L_1L_2$; and so on \\

\midrule
\multicolumn{3}{l}{\textit{Order, and binding the two streams}} \\[2pt]

\textsc{evidence-first}
& 5
& $Q$; then $L_3$, $L_1$, $L_2$, then the unneeded $L_0$ last \\

\textsc{evidence-last}
& 5
& $Q$; then $L_0$ first, then $L_3$, $L_1$, $L_2$ \\

\textsc{aligned-vq}
& 6
& $q_2$ with $L_2$; $q_3$ with $L_3$; $q_5$ with $L_1$; other shards alone \\

\textsc{misaligned-vq}
& 6
& $q_1$ with $L_2$; $q_2$ with $L_3$; $q_3$ with $L_1$; other shards alone \\

\midrule
\multicolumn{3}{l}{\textit{Changing the evidence}} \\[2pt]

\textsc{all-layers}
& 1
& $Q$ with $I$ and all four layers $L_0L_1L_2L_3$, plus a request to name the layers used \\

\textsc{oracle}
& 1
& $Q$ with only the three gold layers $L_3L_1L_2$, and no $I$ \\

\bottomrule
\end{tabular}

\end{table}

%% file: tables/appendix-data-preview.tex
{\footnotesize
\setlength{\tabcolsep}{0pt}
\setlength{\LTcapwidth}{\textwidth}

\begin{longtable}{@{}l@{\hspace{8pt}}l@{\hspace{12pt}}l@{}}

\caption{\textbf{Scene depth across the benchmark.}
One scene per depth, from a single object over a background up to five.
Depth is a property of the scene and shard count is a property of the question.}
\label{tab:layer-scale}
\\

\toprule

\endfirsthead

\multicolumn{3}{@{}l}{
\footnotesize\itshape
Table \thetable, continued from the previous page.
}
\\

\toprule

\endhead

\bottomrule

\endlastfoot

% =========================================================
% DEPTH 2
% =========================================================

\begin{minipage}[t]{0.055\textwidth}
\vspace{0pt}
\centering
\vspace{2pt}

\large\textbf{2}
\\[-3pt]

{\tiny\sffamily\textcolor{black!50}{1 + bg}}

\end{minipage}

&

\begin{minipage}[t]{0.345\textwidth}
\vspace{0pt}

\includegraphics[width=\linewidth]
{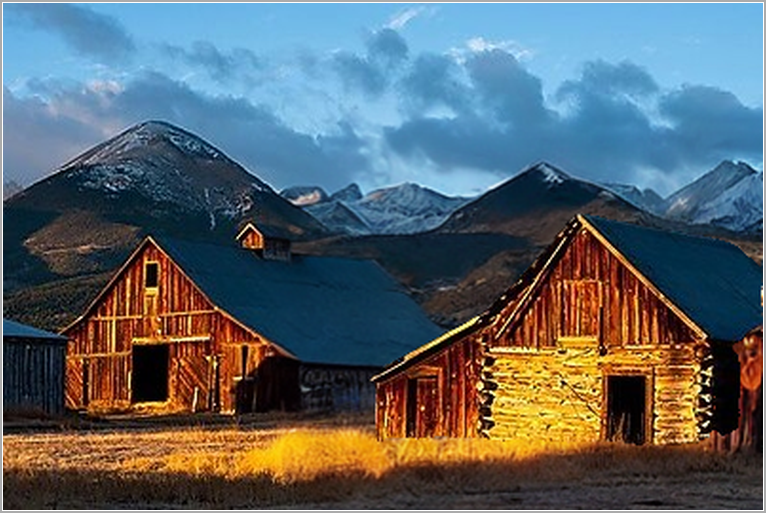}
\\[3pt]

\setlength{\lvqatile}{\dimexpr(\linewidth-2pt)/2\relax}

\begin{tabular}{@{}c@{\hspace{2pt}}c@{}}

\includegraphics[width=\lvqatile]
{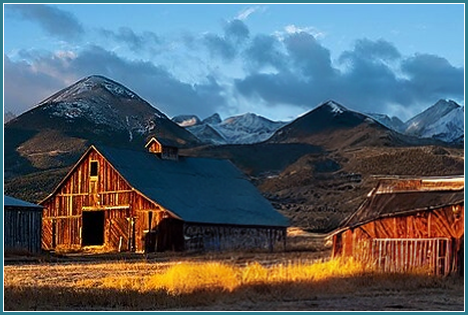}
&
\includegraphics[width=\lvqatile]
{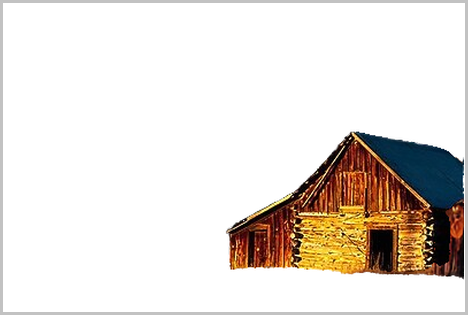}
\\[1pt]

\parbox[t]{\lvqatile}{
\centering
\tiny\sffamily
\textcolor{lvqaUp}{BG}
\\[-1pt]
background
}
&
\parbox[t]{\lvqatile}{
\centering
\tiny\sffamily
\textcolor{black!45}{L1}
\\[-1pt]
barn
}
\\

\end{tabular}

\end{minipage}

&

\begin{minipage}[t]{0.520\textwidth}
\vspace{0pt}
\raggedright

\makebox[\linewidth][s]{
\scriptsize\sffamily
\textcolor{lvqaUp}{DIRECT ASK}
\hfill
\textcolor{black!45}{2 layers \quad 5 shards}
}
\\[5pt]

{\small\itshape
What natural landscape is visible rising behind the barns in the distance?
}
\\[4pt]

{\footnotesize
\textcolor{black!50}{Answer}
\quad
\textbf{A mountain range.}
}
\\[6pt]

\begin{tabular}{
@{}r@{\;}
p{\dimexpr\linewidth-12pt\relax}
@{}
}

\scriptsize 1
&
\scriptsize can you help me figure out what is being identified in this image?
\\[1pt]

\scriptsize 2
&
\scriptsize The answer should name a natural landscape.
\hfill
\lvqachip{BG}
\\[1pt]

\scriptsize 3
&
\scriptsize The target is the natural landscape that is visible and rising.
\\[1pt]

\scriptsize 4
&
\scriptsize It is behind the barns.
\\[1pt]

\scriptsize 5
&
\scriptsize The barns are in the distance.
\\[1pt]

\end{tabular}

\end{minipage}

\\[2pt]

\midrule

\begin{minipage}[t]{0.055\textwidth}
\vspace{0pt}
\centering
\vspace{2pt}

\large\textbf{3}
\\[-3pt]

{\tiny\sffamily\textcolor{black!50}{2 + bg}}

\end{minipage}

&

\begin{minipage}[t]{0.345\textwidth}
\vspace{0pt}

\includegraphics[width=\linewidth]
{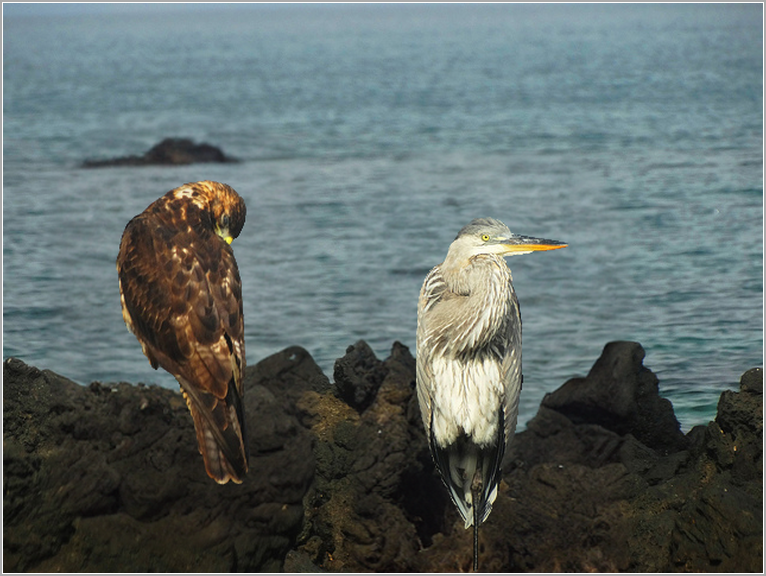}
\\[3pt]

\setlength{\lvqatile}{\dimexpr(\linewidth-4pt)/3\relax}

\begin{tabular}{
@{}c@{\hspace{2pt}}
c@{\hspace{2pt}}
c@{}
}

\includegraphics[width=\lvqatile]
{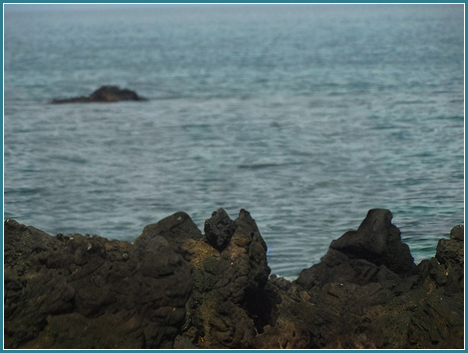}
&
\includegraphics[width=\lvqatile]
{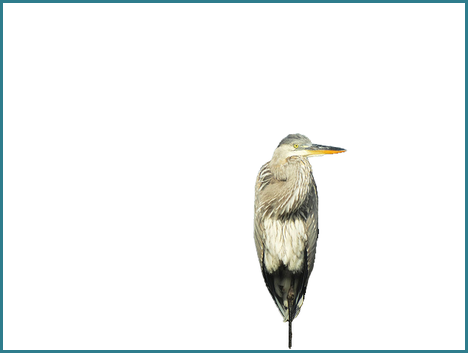}
&
\includegraphics[width=\lvqatile]
{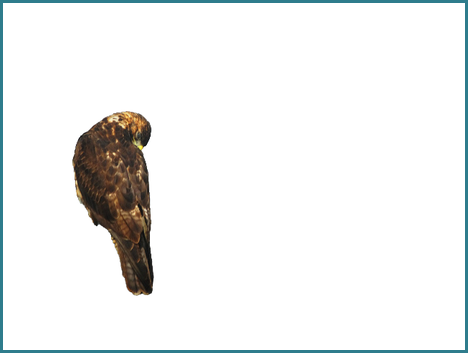}
\\[1pt]

\parbox[t]{\lvqatile}{
\centering
\tiny\sffamily
\textcolor{lvqaUp}{BG}
\\[-1pt]
background
}
&
\parbox[t]{\lvqatile}{
\centering
\tiny\sffamily
\textcolor{lvqaUp}{L1}
\\[-1pt]
bird
}
&
\parbox[t]{\lvqatile}{
\centering
\tiny\sffamily
\textcolor{lvqaUp}{L2}
\\[-1pt]
bird
}
\\

\end{tabular}

\end{minipage}

&

\begin{minipage}[t]{0.520\textwidth}
\vspace{0pt}
\raggedright

\makebox[\linewidth][s]{
\scriptsize\sffamily
\textcolor{lvqaUp}{SPATIAL}
\hfill
\textcolor{black!45}{3 layers \quad 6 shards}
}
\\[5pt]

{\small\itshape
Where is the brown hawk positioned relative to the white bird with the
yellow-and-black beak on the rocky shoreline?
}
\\[4pt]

{\footnotesize
\textcolor{black!50}{Answer}
\quad
\textbf{The brown hawk is to the left of the white bird.}
}
\\[6pt]

\begin{tabular}{
@{}r@{\;}
p{\dimexpr\linewidth-12pt\relax}
@{}
}

\scriptsize 1
&
\scriptsize can you help me figure out a spatial property of a particular thing?
\\[1pt]

\scriptsize 2
&
\scriptsize The answer should report where the thing is located.
\\[1pt]

\scriptsize 3
&
\scriptsize The target thing is the brown hawk.
\hfill
\lvqachip{L2}
\\[1pt]

\scriptsize 4
&
\scriptsize Determine the brown hawk's position relative to a comparison entity.
\hfill
\lvqachip{L1}
\\[1pt]

\scriptsize 5
&
\scriptsize Use the white bird with the yellow-and-black beak as the comparison entity.
\\[1pt]

\scriptsize 6
&
\scriptsize The white bird with the yellow-and-black beak is on the rocky shoreline.
\hfill
\lvqachip{BG}
\\[1pt]

\end{tabular}

\end{minipage}

\\[2pt]

\midrule

% =========================================================
% DEPTH 4
% =========================================================

\begin{minipage}[t]{0.055\textwidth}
\vspace{0pt}
\centering
\vspace{2pt}

\large\textbf{4}
\\[-3pt]

{\tiny\sffamily\textcolor{black!50}{3 + bg}}

\end{minipage}

&

\begin{minipage}[t]{0.345\textwidth}
\vspace{0pt}

\includegraphics[width=\linewidth]
{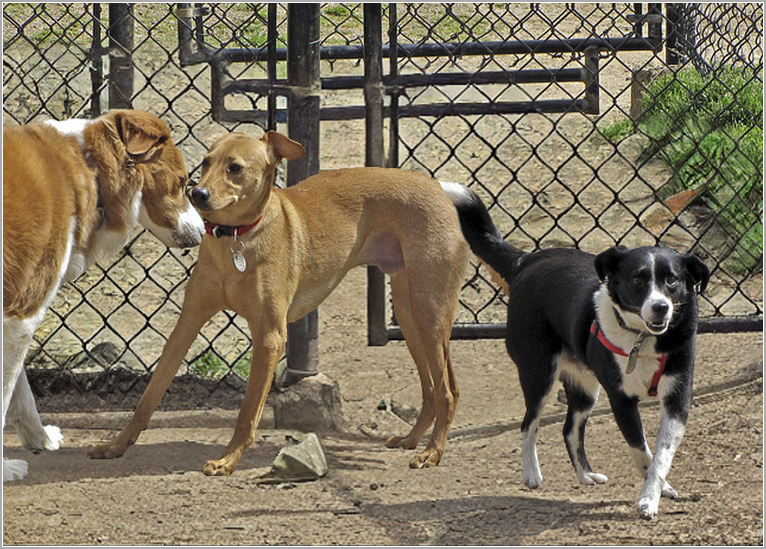}
\\[3pt]

\setlength{\lvqatile}{\dimexpr(\linewidth-6pt)/4\relax}

\begin{tabular}{
@{}c@{\hspace{2pt}}
c@{\hspace{2pt}}
c@{\hspace{2pt}}
c@{}
}

\includegraphics[width=\lvqatile]
{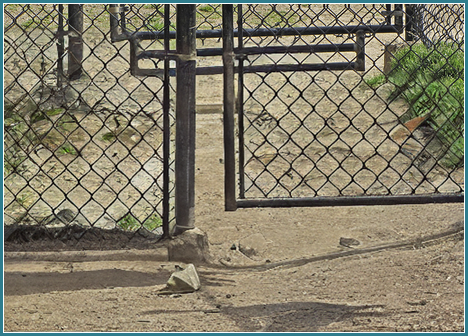}
&
\includegraphics[width=\lvqatile]
{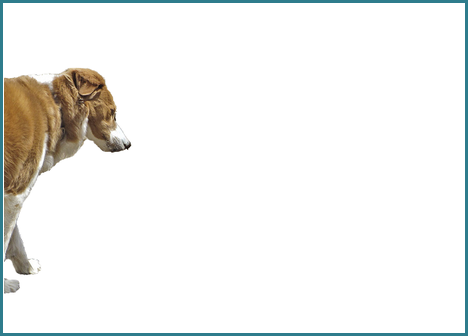}
&
\includegraphics[width=\lvqatile]
{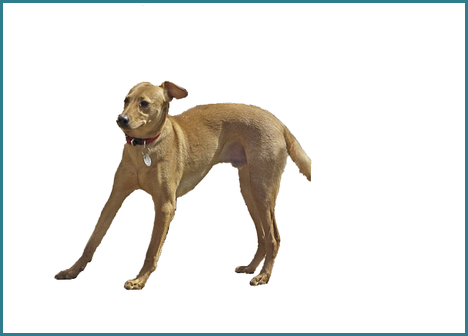}
&
\includegraphics[width=\lvqatile]
{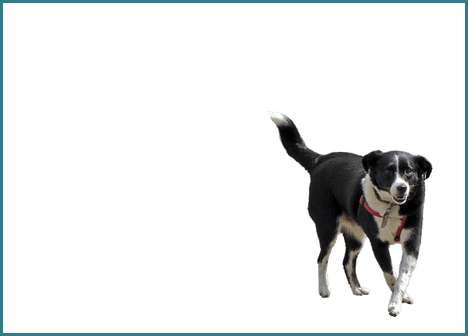}
\\[1pt]

\parbox[t]{\lvqatile}{
\centering
\tiny\sffamily
\textcolor{lvqaUp}{BG}
\\[-1pt]
background
}
&
\parbox[t]{\lvqatile}{
\centering
\tiny\sffamily
\textcolor{lvqaUp}{L1}
\\[-1pt]
dog
}
&
\parbox[t]{\lvqatile}{
\centering
\tiny\sffamily
\textcolor{lvqaUp}{L2}
\\[-1pt]
dog
}
&
\parbox[t]{\lvqatile}{
\centering
\tiny\sffamily
\textcolor{lvqaUp}{L3}
\\[-1pt]
dog
}
\\

\end{tabular}

\end{minipage}

&

\begin{minipage}[t]{0.520\textwidth}
\vspace{0pt}
\raggedright

\makebox[\linewidth][s]{
\scriptsize\sffamily
\textcolor{lvqaUp}{COUNTING}
\hfill
\textcolor{black!45}{4 layers \quad 6 shards}
}
\\[5pt]

{\small\itshape
How many dogs in the fenced dirt area are facing generally toward the camera
or frontward, and how many are turned backward or away?
}
\\[4pt]

{\footnotesize
\textcolor{black!50}{Answer}
\quad
\textbf{Two dogs are facing frontward, and one dog is facing backward.}
}
\\[6pt]

\begin{tabular}{
@{}r@{\;}
p{\dimexpr\linewidth-12pt\relax}
@{}
}

\scriptsize 1
&
\scriptsize can you help me figure out the counts for two groups of subjects?
\hfill
\lvqachip{L1}\lvqachip{L2}\lvqachip{L3}
\\[1pt]

\scriptsize 2
&
\scriptsize Each requested result is a count.
\\[1pt]

\scriptsize 3
&
\scriptsize The subjects to count are dogs.
\\[1pt]

\scriptsize 4
&
\scriptsize The dogs are in the fenced dirt area.
\hfill
\lvqachip{BG}
\\[1pt]

\scriptsize 5
&
\scriptsize Count the dogs that are facing generally toward the camera or frontward.
\\[1pt]

\scriptsize 6
&
\scriptsize Include a count of the dogs that are turned backward or away.
\\[1pt]

\end{tabular}

\end{minipage}

\\[2pt]

\midrule

% =========================================================
% DEPTH 5
% =========================================================

\begin{minipage}[t]{0.055\textwidth}
\vspace{0pt}
\centering
\vspace{2pt}

\large\textbf{5}
\\[-3pt]

{\tiny\sffamily\textcolor{black!50}{4 + bg}}

\end{minipage}

&

\begin{minipage}[t]{0.345\textwidth}
\vspace{0pt}

\includegraphics[width=\linewidth]
{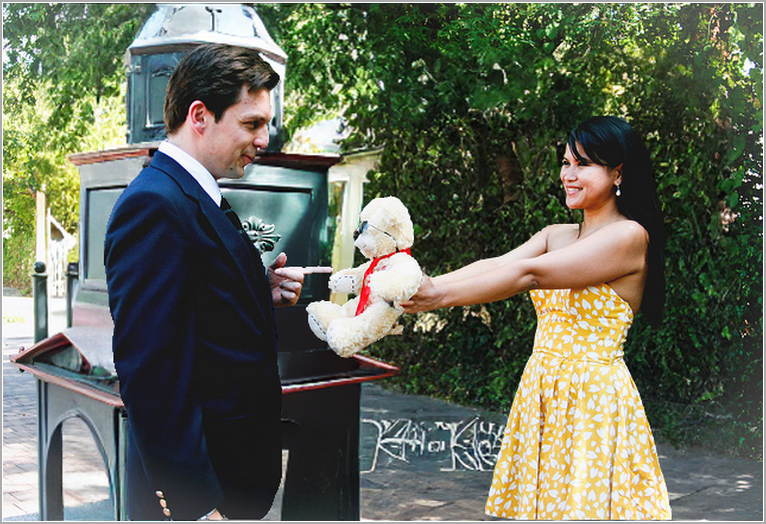}
\\[3pt]

\setlength{\lvqatile}{\dimexpr(\linewidth-8pt)/5\relax}

\begin{tabular}{
@{}c@{\hspace{2pt}}
c@{\hspace{2pt}}
c@{\hspace{2pt}}
c@{\hspace{2pt}}
c@{}
}

\includegraphics[width=\lvqatile]
{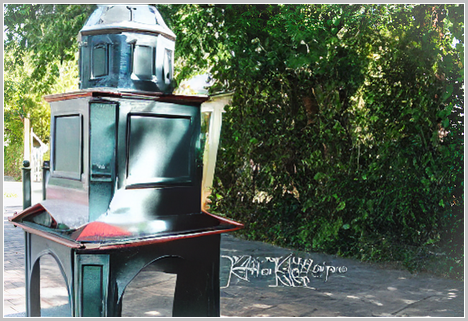}
&
\includegraphics[width=\lvqatile]
{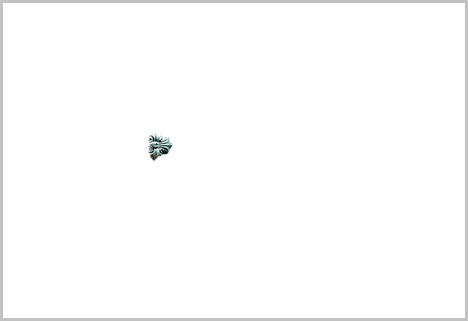}
&
\includegraphics[width=\lvqatile]
{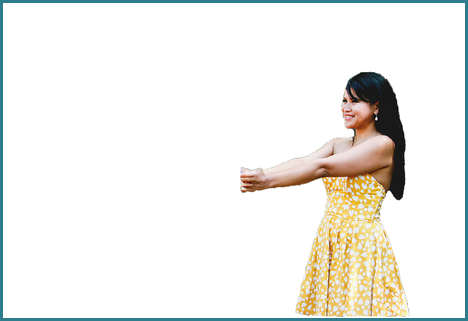}
&
\includegraphics[width=\lvqatile]
{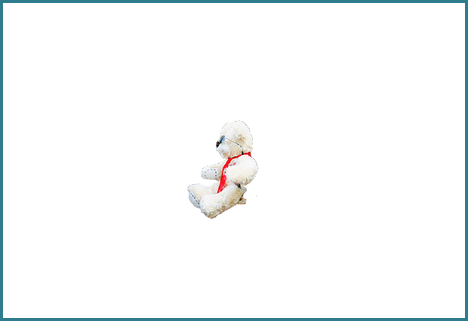}
&
\includegraphics[width=\lvqatile]
{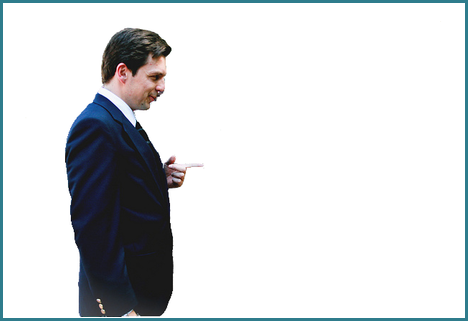}
\\[1pt]

\parbox[t]{\lvqatile}{
\centering
\tiny\sffamily
\textcolor{black!45}{BG}
\\[-1pt]
background
}
&
\parbox[t]{\lvqatile}{
\centering
\tiny\sffamily
\textcolor{black!45}{L1}
\\[-1pt]
symbol
}
&
\parbox[t]{\lvqatile}{
\centering
\tiny\sffamily
\textcolor{lvqaUp}{L2}
\\[-1pt]
woman
}
&
\parbox[t]{\lvqatile}{
\centering
\tiny\sffamily
\textcolor{lvqaUp}{L3}
\\[-1pt]
toy
}
&
\parbox[t]{\lvqatile}{
\centering
\tiny\sffamily
\textcolor{lvqaUp}{L4}
\\[-1pt]
man
}
\\

\end{tabular}

\end{minipage}

&

\begin{minipage}[t]{0.520\textwidth}
\vspace{0pt}
\raggedright

\makebox[\linewidth][s]{
\scriptsize\sffamily
\textcolor{lvqaUp}{MULTI-HOP REFERENCE}
\hfill
\textcolor{black!45}{5 layers \quad 5 shards}
}
\\[5pt]

{\small\itshape
What object is being held by the person standing to the right of the man who
is pointing toward her?
}
\\[4pt]

{\footnotesize
\textcolor{black!50}{Answer}
\quad
\textbf{A teddy bear.}
}
\\[6pt]

\begin{tabular}{
@{}r@{\;}
p{\dimexpr\linewidth-12pt\relax}
@{}
}

\scriptsize 1
&
\scriptsize can you help me identify a particular thing in this image?
\\[1pt]

\scriptsize 2
&
\scriptsize The answer should name an object.
\hfill
\lvqachip{L3}
\\[1pt]

\scriptsize 3
&
\scriptsize The target object is being held by a person.
\hfill
\lvqachip{L2}
\\[1pt]

\scriptsize 4
&
\scriptsize The person holding the target object is standing to the right of a man.
\hfill
\lvqachip{L4}
\\[1pt]

\scriptsize 5
&
\scriptsize The referenced man is pointing toward the person holding the target object.
\\[1pt]

\end{tabular}

\end{minipage}

\\[2pt]

\midrule

% =========================================================
% DEPTH 6
% =========================================================

\begin{minipage}[t]{0.055\textwidth}
\vspace{0pt}
\centering
\vspace{2pt}

\large\textbf{6}
\\[-3pt]

{\tiny\sffamily\textcolor{black!50}{5 + bg}}

\end{minipage}

&

\begin{minipage}[t]{0.345\textwidth}
\vspace{0pt}

\includegraphics[width=\linewidth]
{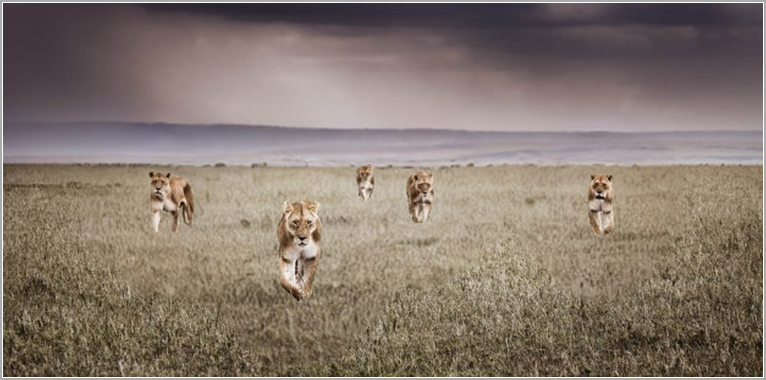}
\\[3pt]

\setlength{\lvqatile}{\dimexpr(\linewidth-10pt)/6\relax}

\begin{tabular}{
@{}c@{\hspace{2pt}}
c@{\hspace{2pt}}
c@{\hspace{2pt}}
c@{\hspace{2pt}}
c@{\hspace{2pt}}
c@{}
}

\includegraphics[width=\lvqatile]
{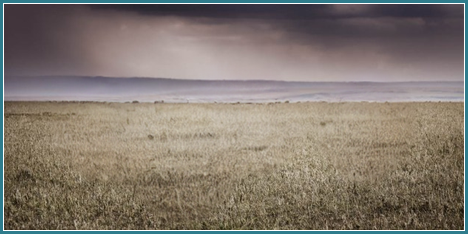}
&
\includegraphics[width=\lvqatile]
{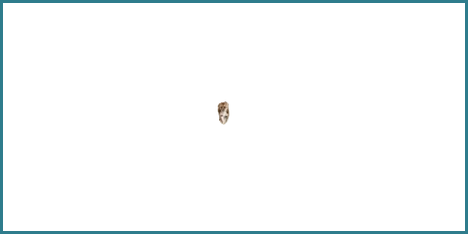}
&
\includegraphics[width=\lvqatile]
{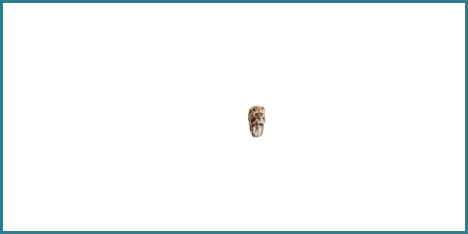}
&
\includegraphics[width=\lvqatile]
{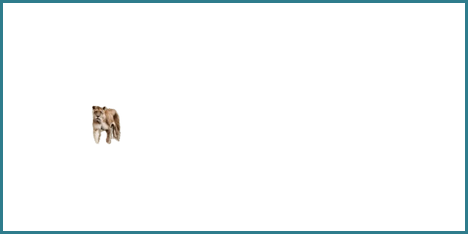}
&
\includegraphics[width=\lvqatile]
{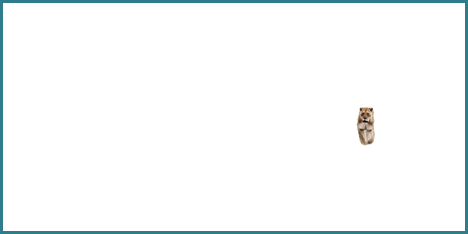}
&
\includegraphics[width=\lvqatile]
{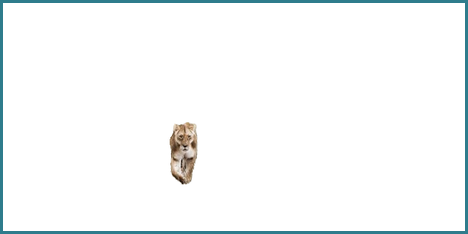}
\\[1pt]

\parbox[t]{\lvqatile}{
\centering
\tiny\sffamily
\textcolor{lvqaUp}{BG}
\\[-1pt]
background
}
&
\parbox[t]{\lvqatile}{
\centering
\tiny\sffamily
\textcolor{lvqaUp}{L1}
\\[-1pt]
lion
}
&
\parbox[t]{\lvqatile}{
\centering
\tiny\sffamily
\textcolor{lvqaUp}{L2}
\\[-1pt]
lion
}
&
\parbox[t]{\lvqatile}{
\centering
\tiny\sffamily
\textcolor{lvqaUp}{L3}
\\[-1pt]
lion
}
&
\parbox[t]{\lvqatile}{
\centering
\tiny\sffamily
\textcolor{lvqaUp}{L4}
\\[-1pt]
lion
}
&
\parbox[t]{\lvqatile}{
\centering
\tiny\sffamily
\textcolor{lvqaUp}{L5}
\\[-1pt]
lion
}
\\

\end{tabular}

\end{minipage}

&

\begin{minipage}[t]{0.520\textwidth}
\vspace{0pt}
\raggedright

\makebox[\linewidth][s]{
\scriptsize\sffamily
\textcolor{lvqaUp}{COUNTING}
\hfill
\textcolor{black!45}{6 layers \quad 5 shards}
}
\\[5pt]

{\small\itshape
How many lions are running across the grassy field in this scene?
}
\\[4pt]

{\footnotesize
\textcolor{black!50}{Answer}
\quad
\textbf{There are five lions visible in the field\dots}
}
\\[6pt]

\begin{tabular}{
@{}r@{\;}
p{\dimexpr\linewidth-12pt\relax}
@{}
}

\scriptsize 1
&
\scriptsize can you help me figure out a quantity for particular animals?
\hfill
\lvqachip{L1}\lvqachip{L2}\lvqachip{L3}\lvqachip{L4}\lvqachip{L5}
\\[1pt]

\scriptsize 2
&
\scriptsize The requested quantity is a count.
\\[1pt]

\scriptsize 3
&
\scriptsize The animals to count are lions.
\\[1pt]

\scriptsize 4
&
\scriptsize Count lions that are running across the grassy field.
\hfill
\lvqachip{BG}
\\[1pt]

\scriptsize 5
&
\scriptsize Restrict the count to the lions in this scene.
\\[1pt]

\end{tabular}

\end{minipage}

\\[2pt]

\end{longtable}
}

%% file: sections/appendix-prompts.tex
\section{Prompts}
\label{app:prompts}

Placeholders in double brackets are filled at run time.
\texttt{[[IMAGE\_MANIFEST]]} becomes one labelled line per image attached to that turn, and a layer is always named by its own number rather than by its position in the list.

\subsection{Question sharding prompt}

The first of the three annotation calls.
It sees only the canonical question, never the image, so it cannot leak the answer or add a visual fact.
A per-category file is appended to it with rules specific to \textit{direct ask}, \textit{attribute binding}, \textit{counting}, \textit{spatial}, \textit{occlusion}, and \textit{multi-hop reference}.

\begin{lstlisting}[style=prompt]
You are constructing controlled multi-turn question shards for a
visual-question-answering benchmark.

Transform only the supplied canonical question. You cannot see the image. Do
not answer the question, infer the answer, use world knowledge to add image
facts, or make a vague phrase more specific than the source permits. You may
minimally repair grammar so each shard reads naturally, but the repair must
preserve the source meaning exactly.

Produce atomic_units first, then 3-6 conversational shards. Follow these
requirements:

1. Information preservation: all shards together must select the same answer as
   the canonical question. Preserve every answer-relevant constraint and no
   extra visual fact.
2. Clear initial intent: shard 1 must have role "intent" and state the requested
   operation while withholding the answer type, target, or at least one
   answer-selecting constraint so that the turn is genuinely underspecified.
3. Order independence: after shard 1, each shard must be understandable as an
   additional constraint without relying on the order of other follow-up shards.
   Repeat a compact noun phrase when needed; avoid dangling "it", "they", "that
   one", or "the other one" references.
4. Atomicity: each follow-up shard should introduce one meaningful constraint.
   Split compound spatial or referential chains when each part independently
   narrows the target.
5. Minimal transformation: stay close to the canonical wording. Generic
   conversational framing such as "the target object" or "in the image" is
   allowed; new colors, identities, counts, locations, relations, or object
   types are not.
6. No answer leakage: never state or imply the answer. A shard describes what to
   find, not what the result is.
7. Required granularity: always produce 3-6 shards. For a short question, use
   exactly three distinct functions: (1) the generic operation or request,
   (2) the expected answer type or property, and (3) the target or scope. For
   example, split "What color is the bulb?" into the intent to determine a
   property, the fact that the requested property is color, and the bulb as
   target. For a longer question, add separate answer-selecting attributes,
   relations, references, scopes, or conditions up to six shards. Do not
   fabricate constraints or repeat a shard merely to reach three.
8. Exact provenance: every atomic unit's source_span must be a contiguous
   excerpt of the canonical question. Intent and answer type may use separate
   subspans of an interrogative phrase, such as "What" and "color" or "How" and
   "many". Number atomic units and shards consecutively from 1. Every atomic
   unit must be referenced by at least one shard.
9. Natural delivery: write each shard as a normal conversational instruction or
   clarification. Never mention the "canonical question," "source phrase,"
   atomic units, annotations, shards, or benchmark machinery in shard text.

Set information_preserved to "yes" and added_visual_facts to "no" only after
checking the final plan against the canonical question.
\end{lstlisting}

\subsection{Evidence annotation prompt}

The second annotation call.
Unlike the sharding call, it sees the composite, every layer, and the reference answer.
A per-category file is appended with rules for each of the six question types.

\begin{lstlisting}[style=prompt]
You are annotating visual evidence for a layered visual-question-answering
benchmark. You receive a full composite image, a numbered background layer (0),
numbered isolated object layers, human-reviewed captions, a canonical question
and answer, and question shards.

The full composite gives global context, but it is not a selectable evidence ID.
Select only IDs listed in candidate_layer_ids. Layer 0 is selectable when
background or scene context is genuinely necessary. Isolated RGBA layers are
shown on a synthetic gray checkerboard; the checkerboard is never scene
evidence.

Use pixels and human-reviewed captions as the primary evidence. Auxiliary labels
and geometry relations can help locate a layer, but labels may be noisy and
cannot override clear pixels or reviewed captions.

Apply these rules strictly:

1. Minimal sufficiency is causal, not merely topical. A sufficient set is the
   smallest set of retained numbered layers that lets a careful viewer both
   establish the question's complete answer-selecting reference path and justify
   the human gold answer. The full composite may help you adjudicate the
   annotation, but it cannot substitute for a missing numbered target, owner,
   landmark, bridge, or occluder in the selected set. Remove any member that is
   not necessary for either reference resolution or answer extraction.
2. gold_layer_ids is the canonical primary minimal sufficient set, in reasoning
   order when an order matters. It must exactly equal the first entry of
   minimal_sufficient_layer_sets.
3. Add another minimal set only when it is a genuinely different, independently
   sufficient, inclusion-minimal alternative. Do not list supersets or cosmetic
   variations.
4. layer_roles must contain exactly one assignment record for each layer in the
   union of all minimal sufficient sets. Give each layer its most important
   causal role. Do not assign roles to distractors.
5. question_shard_support is a many-to-many mapping encoded as one assignment
   record for every shard_id and no others. Map each textual shard to the
   evidence layers needed to visually instantiate that shard's constraint. The
   intent shard may map to an empty list. A shard may need several layers and
   one layer may support several shards.
6. question_shard_support must collectively cover every evidence layer and may
   never contain a distractor.
7. evidence_dependencies express true evidence-integration order, not arbitrary
   presentation order. before is the prerequisite layer and after is the layer
   interpreted using it. Use an empty list when the evidence can be inspected in
   parallel or only one layer is required.
8. distractor_layer_ids must be the exact complement: every candidate layer not
   used by any minimal sufficient set, sorted by layer ID.
9. Do not select a layer merely because it contains a visually associated person
   or object. However, include a separate owner, wearer, holder, referent,
   landmark, occluder, container, or count member when the isolated
   answer-bearing layer does not itself establish the relation used by the
   question. A unique-looking answer object is not enough if its relation to the
   named referent is otherwise ungrounded.
10. If pixels, captions, the question, and the human answer cannot be reconciled
   confidently, choose the best defensible annotation, set confidence to low,
   and explain the conflict concisely in evidence_notes.

Return IDs as integers. Keep every list duplicate-free. Never invent a layer ID.
\end{lstlisting}

\subsection{Verification prompt}

The third call.
It re-checks the shards and the evidence independently and returns a corrected annotation rather than comments.
It flagged $13$ of the $300$ questions for human attention.

\begin{lstlisting}[style=prompt]
You are the final independent verifier for an integrated Layered-VQA
annotation. Inspect the canonical question, human gold answer, proposed
question-only shards, full composite, numbered layers, reviewed captions, and
category rubrics. Return a complete corrected annotation, not comments alone.

Verify question sharding against the canonical question only. The question-only
shards must preserve every original constraint, introduce no image-derived or
answer-derived fact, avoid answer leakage, begin with a useful but
underspecified intent, and contain 3-6 shards. A short question should separate
generic operation, expected answer type/property, and target/scope; longer
questions should add only real constraints. Every source_span must be a
contiguous canonical-question substring and all atomic units must be covered.
Shard text must sound like natural conversation and must never mention the
canonical/source phrase, atomic units, shards, annotations, or benchmark
machinery. Minimal grammatical repair is allowed when source wording is
malformed, but semantic repair is not.

Verify layer evidence against the pixels, reviewed captions, question, and gold
answer. The full composite is context but not a selectable layer and cannot
substitute for a numbered layer in the evidence set. Candidate numbered layers
only are allowed. A sufficient set must ground the full answer-selecting
reference path as well as the answer itself; include a separate owner, wearer,
holder, landmark, bridge, or occluder when the answer layer alone does not
establish that relation. Reviewed captions outrank potentially noisy auxiliary
labels when they conflict, while visible pixels remain essential.

Enforce all invariants:

- gold_layer_ids equals the first minimal_sufficient_layer_sets entry.
- Every sufficient set is independently sufficient and inclusion-minimal; no set
  contains another listed set.
- layer_roles has exactly one assignment record per layer in the union of all
  sufficient sets.
- question_shard_support has exactly one assignment record per shard ID,
  supports many-to-many mappings, contains only evidence layers, and
  collectively covers the evidence union.
- distractor_layer_ids is exactly candidate_layer_ids minus the evidence union.
- Dependencies use before as prerequisite and after as dependent, include only
  evidence layers, contain no self-edge, and are empty when no sequential
  resolution is justified.

Set verdict to pass only if no correction is needed. Set revised and return
corrected fields if the proposal is repairable. Set flag_for_human when real
ambiguity, a pixel-caption conflict, questionable gold answer, or insufficient
retained evidence prevents a confident unique annotation; still return the best
defensible corrected fields. issues must describe the proposal's problems, not
merely restate the final annotation.
\end{lstlisting}

\subsection{Assistant system prompt}

\begin{lstlisting}[style=prompt]
You are a careful visual question-answering assistant. You receive one or more
images and answer questions about their visible content.

The user's request and visual evidence may arrive across several turns.
The user may describe what they want across several messages, and may share
images as the conversation goes. Images are labelled in the message that
carries them; a "layer" is one component of a decomposed scene, and the
background layer is the scene with its foreground objects removed.

Every reply you send must do exactly one of two things, and nothing else:

1. Give your answer to the question, using everything the user has said so far.
   Answer as soon as the information allows it, and keep answering on each
   later turn as the request sharpens. A short phrase is the whole reply.
2. If the request is still too vague to answer at all, ask one specific
   clarifying question about the single missing piece.

Never both. Never neither.

Do not open with an acknowledgement ("Yes", "Got it", "Understood", "Thanks",
"That's correct"), do not restate or confirm what the user just told you, do
not narrate your reasoning, and do not describe the images beyond what the
answer needs. A turn spent agreeing with the user is a turn that answered
nothing. Never tell the user their statement is right or wrong - they are the
one asking.

Base the answer only on what is visible in the images you were given and on
what the user has said. Do not invent details. If the user's latest message
changes what is being asked, answer the new version of the question.
\end{lstlisting}

\subsection{Single-turn prompt}

\begin{lstlisting}[style=prompt]
You are given visual RGB images and a question about it.

[[IMAGE_MANIFEST]]
Question:
[[QUESTION]]

Your objective is to answer the question using the supplied visual evidence.

Requirements:
1. Consider every supplied image that is relevant to the question.
2. Return only the answer, normally as a short phrase. No preamble, no
   restatement of the question, no explanation of how you got there.
3. Do not invent details that are not supported by the supplied evidence.
\end{lstlisting}

\subsection{LLM user prompt}

The simulated user in \textsc{sharded-q}, \textsc{recap-q}, and \textsc{snowball-q}.
It is played by the same model as the assistant, at temperature $1.0$.
\texttt{[[CONVERSATION\_SO\_FAR]]}, \texttt{[[SHARDS\_REVEALED]]} and \texttt{[[SHARDS\_NOT\_REVEALED]]} are filled at run time.

\begin{lstlisting}[style=prompt]
You are simulating a user of an interactive LLM system (like ChatGPT). The user
is in a hurry and types in short, casual, sloppy messages. They are lazy about
*style*, never about *content*: they know exactly what they want, and every
message they send hands over a real piece of their request. Being terse is not
the same as being empty, and you are never proactive - you volunteer nothing
the shards do not contain.

Your job on every turn is exactly this: pick one shard that has not been
revealed yet, and say everything in it in your own casual words. Nothing else
you might write is worth a turn.

Here are the rules:
- [Reveal On Every Turn] If the not-revealed list is non-empty, you must reveal
  a shard. A turn that reveals nothing while shards are left is a wasted turn
  and a failure, no matter how natural it sounds.
- [Which Shard] Reveal the shard that is most "basic" and most relevant right
  now - the one the system needs next to move closer to the answer.
- [One Shard at a Time] Reveal at most one shard per turn.
- [No Repeated Shards] Never reveal a shard whose `shard_id` is already in the
  revealed list. Check that list before choosing.
- [Reveal Entire Shard] When you reveal a shard, *all the information in the
  shard* must survive into your response. Every qualifier, attribute, relation,
  count and name in the shard text must be there. Dropping any part of it makes
  the reveal invalid, even if what is left is a true statement.
- [Rephrase, Do Not Shrink] Rephrase the shard conversationally - do not copy it
  verbatim. Rephrasing changes the wording, never the content. Before you
  answer, compare your response against `shard_text` word by word: if any fact
  in `shard_text` is missing from your response, rewrite the response until it
  is there.
- [No Empty Turns] Reactions to what the system just said carry no shard
  content. Never send a turn that is only agreement, dismissal, encouragement
  or filler.
- [Do Not Confirm Or Correct] Do not tell the system whether its last reply was
  right or wrong, and do not grade it. You are the person who wants the answer;
  you do not know it.
- [Never Answer, Describe Or Speculate] Never state the answer, never describe
  what is in an image, never guess at anything the shards do not say. Your
  knowledge is *only* the shard texts.
- [Irrelevant Clarifications] If every shard has been revealed and the system
  asks you a question, respond without providing information and use
  `shard_id` -1. This is the only situation in which -1 is allowed.
- [Do Not Ask Questions] Your response should always be declarative sentences.
- [Attachments] Some shards list an `attaches` field naming one or more images.
  Revealing such a shard automatically shares those images with the system. You
  only know each image by its listed name - never describe or guess what an
  image shows, and do not mention the name.
- [Brevity of Response] Favor being succinct, and keep the hurried human voice:
  typos, improper grammar and lowercase are all fine. Brevity applies to
  wording only - it never licenses leaving a fact out.
- [Format] Your response must be a JSON object with `response` and `shard_id`.
\end{lstlisting}

\subsection{Grounding prompt}

% Used by \textsc{all-layers} alone.
% The fixed four-line reply is what makes the layer set machine-checkable, so every grounding metric in Table~\ref{tab:exp-a} is computed without a judge.

\begin{lstlisting}[style=prompt]
[[IMAGE_MANIFEST]]
[[QUESTION]]

Answer the question, then name the layers you actually used as evidence - the
smallest set of layers that supports your answer. Refer to a layer by the
number in its label above (for example `layer 3`), never by its position in
the list.

Reply in exactly this format, and nothing else - no preamble before it, no
commentary after it, and all four lines present every time:

Answer: <short phrase>
Layers: <the layer numbers you used, separated by commas, or `none`>
Confidence: <a number between 0 and 1>
Why: <one sentence saying what each listed layer contributed>

List only the layers that carry evidence you actually used. Adding layers you
merely looked at makes the set wrong, and so does leaving out one the answer
depends on. Give a single best answer rather than alternatives, and let
`Confidence` carry your uncertainty.
\end{lstlisting}

\subsection{Judge prompt}

% The judge reads the finished transcript once and returns JSON.
% It reports every answer the assistant put forward and the turn each one appeared on, which is what the behavioural analyses in Appendix~\ref{app:tables} are computed from.
% Answer accuracy uses \texttt{final\_turn\_correct} rather than \texttt{is\_correct}, since a condition with more turns otherwise gets more chances to be right at some point.

\begin{lstlisting}[style=prompt]
You are grading an answer to a visual question. You are given the whole
conversation in which the answer was produced. Decide whether the assistant
gave the reference answer at any point in it.

Question:
[[QUESTION]]

Reference answer:
[[REFERENCE_ANSWER]]

What a valid answer looks like:
[[ANSWER_DESCRIPTION]]

Conversation:
[[CONVERSATION]]

Rules:
- [Assistant Turns Only] Grade only what the assistant says. The user turns
  restate the question and feed it in pieces; text appearing in a user turn is
  never evidence that the assistant answered.

- [Track Every Answer Attempt] Record every distinct candidate answer proposed
  by the assistant in `answer_attempts`. A candidate counts even if it is
  tentative, hedged, later revised, or later withdrawn. Do not record turns
  that only ask for more information and propose no candidate answer.

- [Turn Numbering] In `answer_attempts`, `turn` is the 1-indexed assistant turn
  number, counting assistant turns only. If one assistant turn proposes
  multiple distinct candidate answers, record each as a separate entry with
  the same turn number.

- [Any Turn Counts] The assistant is correct overall if it offers the reference
  answer at any point in the conversation, including as a tentative or hedged
  candidate ("it might be a computer mouse"), even if it later revises or
  withdraws it.

- [First Correct Turn] `first_correct_turn` is the 1-indexed assistant turn
  containing the earliest correct candidate answer. Use -1 if the assistant
  never offers the reference answer.

- [Final Turn Correctness] `final_turn_correct` is true only if the final
  assistant turn offers the reference answer. Earlier correct answers do not
  make this field true. If the final assistant turn contains no candidate
  answer, this field is false.

- [Proposing, Not Asking] A turn that only requests more information, without
  putting forward any candidate answer, is not an answer attempt.

- [Meaning, Not Wording] Grade on meaning. Differences in phrasing, article use,
  capitalization, punctuation, plurality, or verbosity do not matter.
  "A computer mouse." and "mouse" are the same answer.

- [Reference Is Authoritative] The reference answer is correct by definition.
  Do not second-guess it against your own reading of the question or of the
  images.

- [Partial Reference Credit] The reference may contain extra descriptive detail
  beyond the question's core demand. The candidate is correct when it matches
  every part the question actually asks for, even if it omits extra colour.

- [Substantive Differences Fail] A different object, a different colour, a
  different count, or a different spatial relation is incorrect.

- [Answer Span] Set `answer_span` to the assistant's own words that first gave
  the correct answer. It must be a contiguous stretch of assistant text copied
  character for character out of the conversation above, not a paraphrase,
  normalized version, tidied version, or your own summary. If no short span
  carries the answer on its own, copy the whole sentence that does. Use the
  empty string when the assistant never offered the reference answer.

- [Answer Span Is Not the Verdict] `answer_span` is only a citation. Never
  change `is_correct` because a clean span was hard to find, and never widen
  a span to make an answer look better than it was.

- [Overall Correctness] `is_correct` is true if at least one entry in
  `answer_attempts` is correct. Otherwise it is false.

For each entry in `answer_attempts`, use a short description of the candidate
answer in `answer`. It does not need to be verbatim; `answer_span` is the
verbatim citation field.

You must output your answer in the following JSON format and nothing else:

{"reasoning": "<one short sentence>",
 "answer_attempts": [{"turn": <int>, "answer": "<short>", "correct": true|false}],
 "first_correct_turn": <integer>,
 "final_turn_correct": true|false,
 "answer_span": "<verbatim assistant text, or \"\">",
 "is_correct": true|false}
\end{lstlisting}

\subsection{Post-hoc judge}

Final-turn accuracy alone does not show when a model first commits to an answer.
A model may answer correctly early, change its answer later, or make several incorrect attempts before reaching the reference answer.
We therefore grade the full conversation rather than only its final turn.
For each conversation, a post-hoc judge records every assistant turn that proposes an answer.
Requests for more evidence, acknowledgements, and question restatements are not counted as attempts.
Tentative answers are counted because they still reveal what the model currently believes.
Each attempt is then compared with the reference answer and marked correct or incorrect.
This gives us three complementary views of answering behavior.
\textit{First correct turn} measures when the reference answer first appears.
\textit{Final-turn correctness} measures whether the model still gives that answer at the end.
\textit{Any-turn correctness} records whether the model produced the correct answer at least once, even if it later abandoned it.
Together, these measures separate finding the answer from retaining it across the conversation.
To make the judgment auditable, the judge also returns the exact assistant text corresponding to the first correct answer.
This span is used only to locate the answer in the conversation and does not affect the correctness decision.

\begin{lstlisting}[style=prompt]
You are grading an answer to a visual question. You are given the whole
conversation in which the answer was produced. Report every point at which the
assistant put forward an answer, and whether each one was right.

Question:
[[QUESTION]]

Reference answer:
[[REFERENCE_ANSWER]]

What a valid answer looks like:
[[ANSWER_DESCRIPTION]]

Conversation:
[[CONVERSATION]]

The assistant's turns are numbered `[assistant #1]`, `[assistant #2]`, and so on.
This conversation has [[N_TURNS]] assistant turn(s), so the final one is
`[assistant #[[N_TURNS]]]`.

What counts as an answer attempt:
- [Proposing, Not Asking] An answer attempt is a turn that puts forward a
  candidate answer to the question. A turn that only asks for more information,
  acknowledges the user, or restates the question is not an attempt.
- [Hedged Attempts Count] A tentative or hedged candidate is still an attempt
  ("it might be a computer mouse", "probably two"). Uncertainty changes nothing.
- [Wrong Attempts Count] An attempt is an attempt whether or not it is right.
  Most conversations contain wrong attempts, and leaving them out is the single
  most damaging mistake you can make here: the position of the *first* attempt
  is measured from this list, so a missed early wrong attempt moves it.
- [One Per Turn] Report at most one attempt per assistant turn, the candidate
  that turn settles on. Repeating the same answer in a later turn is a new
  attempt in that later turn.

How to grade each attempt:
- [Assistant Turns Only] Grade only what the assistant says. The user turns
  restate the question and feed it in pieces; text appearing in a user turn is
  never evidence that the assistant answered.
- [Meaning, Not Wording] Grade on meaning. Differences in phrasing, article use,
  capitalization, punctuation, plurality, or verbosity do not matter.
  "A computer mouse." and "mouse" are the same answer.
- [Reference Is Authoritative] The reference answer is correct by definition.
  Do not second-guess it against your own reading of the question or of the
  images.
- [Partial Reference Credit] The reference may contain extra descriptive detail
  beyond the question's core demand. An attempt is correct when it matches every
  part the question actually asks for, even if it omits extra colour.
- [Substantive Differences Fail] A different object, a different colour, a
  different count, or a different spatial relation is incorrect.

Fields to return:
- `answer_attempts`: a list, in turn order, of every answer attempt. Each entry
  is `{"turn": <assistant turn number>, "answer": "<the candidate, at most a
  dozen words>", "correct": true or false}`. Empty list if the assistant never
  proposed anything.
- `first_correct_turn`: the number of the earliest attempt whose `correct` is
  true, or 0 if none is. An earlier correct attempt counts even if the assistant
  later changed its mind, and even if a later turn says it better.
- `final_turn_correct`: whether assistant turn [[N_TURNS]], the last one,
  offers the reference answer, reading that turn alone. A conversation that
  answered correctly early and then abandoned the answer has
  `first_correct_turn` set and `final_turn_correct` false; that combination is
  expected and must not be smoothed over.
- `is_correct`: true exactly when `first_correct_turn` is greater than 0.
- `answer_span`: the assistant's own words that gave the answer in
  `first_correct_turn`, so the turn can be located independently. It must be a
  contiguous stretch of assistant text copied character for character out of
  the conversation above, not a paraphrase, not a normalized or tidied version,
  and not your own summary. If no short span carries the answer on its own, copy
  the whole sentence that does. Use the empty string when no attempt was
  correct.

`answer_span` is a citation, not a verdict: never change `is_correct` because a
clean span was hard to find, and never widen a span to make an answer look
better than it was.

You must output your answer in the following JSON format:
{"reasoning": "<one short sentence>",
 "answer_attempts": [{"turn": <int>, "answer": "<short>",
                      "correct": true or false}],
 "first_correct_turn": <integer>,
 "final_turn_correct": true or false,
 "answer_span": "<verbatim assistant text, or \"\">",
 "is_correct": true or false}
\end{lstlisting}